\documentclass[10pt,twocolumn,letterpaper]{article}

\usepackage[pagenumbers]{wacv} 

\definecolor{wacvblue}{rgb}{0.21,0.49,0.74}
\usepackage[pagebackref,breaklinks,colorlinks,allcolors=wacvblue]{hyperref}
\usepackage{animate}
\usepackage[utf8]{inputenc} 
\usepackage[T1]{fontenc}    
\usepackage{hyperref}       
\usepackage{url}            
\usepackage{booktabs}       
\usepackage{amsfonts}       
\usepackage{nicefrac}       
\usepackage{microtype}      
\usepackage{xcolor}         
\usepackage{caption}
\usepackage{subcaption}
\usepackage{wrapfig}
\usepackage{multirow} 
\usepackage{paralist}
\usepackage{marvosym}
\usepackage{xspace}
\usepackage{color}
\usepackage{multirow}
\usepackage{ifthen}
\usepackage{amsmath}

\usepackage[most]{tcolorbox}
\usepackage{listings}
\definecolor{lightgray}{gray}{0.95}
\definecolor{deepblue}{RGB}{70,130,180}
\definecolor{deepgray}{RGB}{119,136,153}
\lstdefinestyle{prompt}{
    basicstyle=\ttfamily\fontsize{7pt}{8pt}\selectfont,
    frame=none,
    breaklines=true,
    backgroundcolor=\color{lightgray},
    breakatwhitespace=true,
    breakindent=0pt,
    escapeinside={(*@}{@*)},
    numbers=none,
    numbersep=5pt,
    xleftmargin=5pt,
    aboveskip=2pt,
    belowskip=2pt,
}
\tcbset{
  aibox/.style={
    top=10pt,
    colback=white,
    enhanced,
    center,
  }
}
\newtcolorbox{AIbox}[2][]{aibox, title=#2,#1}

\long\def\ignorethis#1{}

\definecolor{gray}{rgb}{0.35,0.35,0.35}
\definecolor{MyBlue}{rgb}{0,0.2,0.8}
\definecolor{MyRed}{rgb}{0.8,0.2,0}
\definecolor{MyGreen}{rgb}{0.0,0.5,0.1}
\definecolor{MyGray}{rgb}{0.4,0.4,0.4}

\newlength\paramargin
\newlength\figmargin
\newlength\subfigmargin
\newlength\secmargin
\newlength\subsecmargin
\newlength\tabmargin
\newlength\eqmargin

\usepackage{array}
\newcolumntype{L}[1]{>{\raggedright\let\newline\\\arraybackslash\hspace{0pt}}m{#1}}
\newcolumntype{C}[1]{>{\centering\let\newline\\\arraybackslash\hspace{0pt}}m{#1}}
\newcolumntype{R}[1]{>{\raggedleft\let\newline\\\arraybackslash\hspace{0pt}}m{#1}}

\def\ie{i.e.,~}
\def\eg{e.g.,~}

\def\vs{vs.~}

\newcommand{\subsecref}[1]{Section~\ref{subsec:#1}}
\newcommand{\figref}[1]{Fig.~\ref{fig:#1}}
\newcommand{\tabref}[1]{Table~\ref{tab:#1}}

\newcommand{\Paragraph}[1]{\noindent\textbf{#1}}

\title{UniVid: Unifying Vision Tasks with Pre-trained Video Generation Models}

\author{
Lan Chen$^{1}$ \hspace{0.65cm}  Yuchao Gu$^2$ \hspace{0.65cm}  Qi Mao$^{1,\textsuperscript{\Letter}}$ \\
$^1$MIPG, Communication University of China \hspace{0.65cm} $^2$Show Lab, National University of Singapore
}

\begin{document}
\maketitle
\begin{abstract}
Large language models, trained on extensive corpora, successfully unify diverse linguistic tasks within a single generative framework.
Inspired by this, recent works like Large Vision Model (LVM) extend this paradigm to vision by organizing tasks into sequential visual sentences, where visual prompts serve as the context to guide outputs.
However, such modeling requires task-specific pre-training across modalities and sources, which is costly and limits scalability to unseen tasks.
Given that pre-trained video generation models inherently capture temporal sequence dependencies, we explore a more unified and scalable alternative: \textbf{can a pre-trained video generation model adapt to diverse image and video tasks?}
To answer this, we propose \textbf{UniVid}, a framework that fine-tunes a video diffusion transformer to handle various vision tasks without task-specific modifications.
Tasks are represented as visual sentences, where the context sequence defines both the task and the expected output modality.
We evaluate the generalization of UniVid from two perspectives:
(1) \textbf{cross-modal} inference with contexts composed of both images and videos, extending beyond LVM’s uni-modal setting;
(2) \textbf{cross-source} tasks from natural to annotated data, without multi-source pre-training.
Despite being trained solely on natural video data, UniVid generalizes well in both settings.
Notably, understanding and generation tasks can easily switch by simply reversing the visual sentence order in this paradigm.
These findings highlight the potential of pre-trained video generation models to serve as a scalable and unified foundation for vision modeling.
Our code will be released at \url{https://github.com/CUC-MIPG/UniVid}.
\end{abstract}    
\begin{figure*}[ht]
    \centering
\includegraphics[width=1\linewidth]{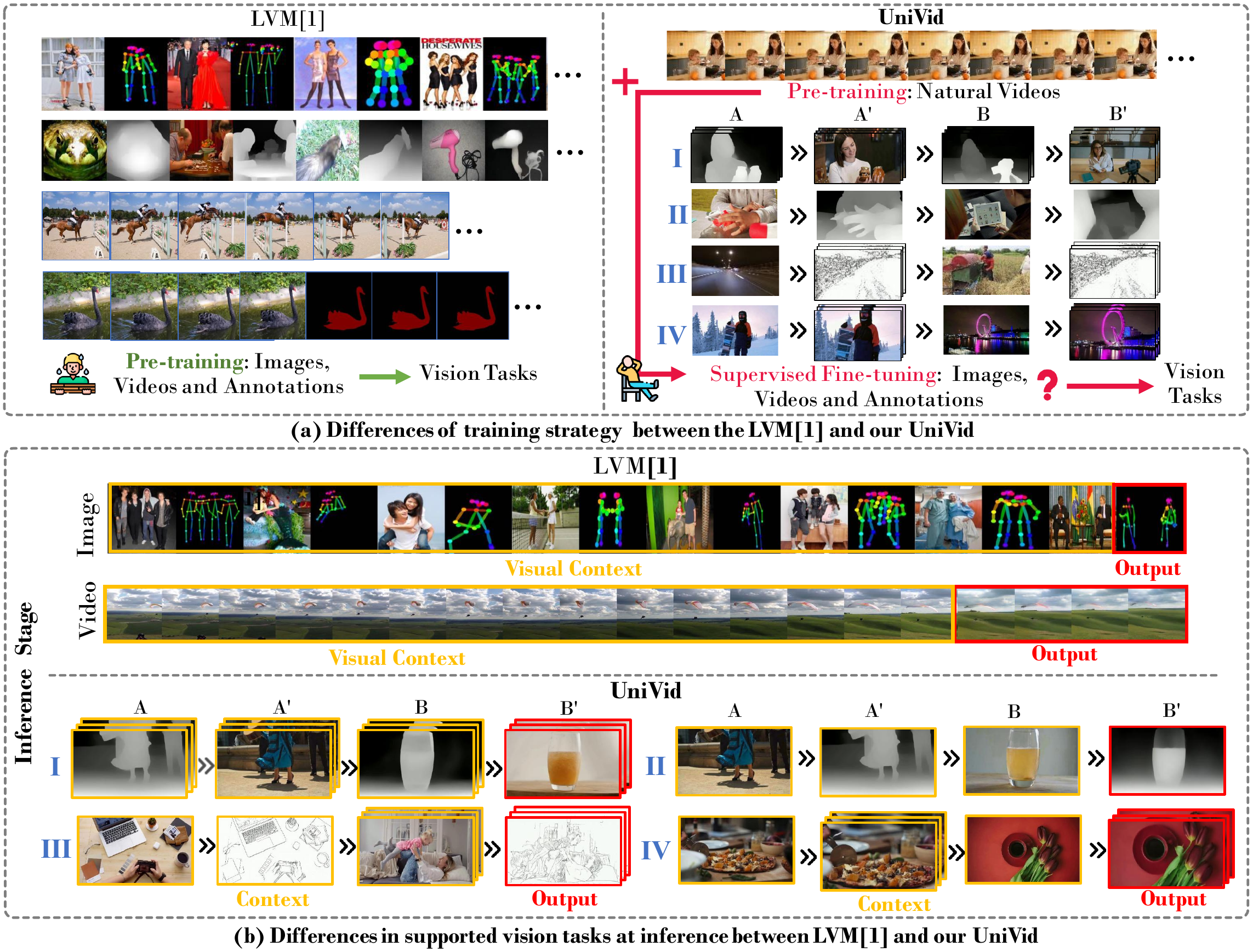}    
\vspace{-6 mm}
  \caption{
\textbf{LVM~\cite{bai2024sequential} \vs UniVid.}
(a) LVM~\cite{bai2024sequential} requires large-scale, modality- and source-specific paired data for pre-training to support diverse vision tasks. In contrast, UniVid explores whether a pre-trained video generation model can be efficiently adapted to a broad range of vision tasks via lightweight SFT with minimal paired data.
(b) At inference, LVM~\cite{bai2024sequential} is limited to uni-modal visual contexts, whereas UniVid enables a unified framework that accommodates both cross-modal and cross-source vision tasks. 
Stacked blocks represent videos; a single block represents an image.
}
    \vspace{-3 mm}
    \label{fig:teaser}
\end{figure*}
\section{Introduction}
Large language models (LLMs) such as GPT~\cite{brown2020language} and DeepSeek-R1~\cite{guo2025deepseek} have garnered significant attention for their ability to tackle a broad spectrum of language tasks within a unified framework.
This success motivates the pursuit of developing similar unified models for various vision tasks.
A representative effort in this direction is the Large Vision Model (LVM)~\cite{bai2024sequential}, which seeks to bridge the gap by representing diverse vision tasks—including both natural data (images, videos) and annotated data (\eg segmentation maps)—as \textit{visual sentences} in pixel space.
LVM~\cite{bai2024sequential} processes these heterogeneous inputs using a unified sequential architecture, analogous to how language models handle natural language sequences.

Within this framework, sequential task-specific data—including images, videos, or annotations—compose the visual prompt (or \textit{visual context}) that guides output generation, as illustrated in ~\figref{teaser}(a).
However, realizing such unified visual modeling in practice remains challenging: LVM~\cite{bai2024sequential} still requires pre-training on separate, task- and modality-specific datasets (\eg for generation \vs understanding, and for images \vs videos).
This fragmented data curation process is highly labor-intensive and fundamentally constrains scalability to new tasks.
These limitations motivate us to explore a more unified and scalable approach for vision modeling.

Unlike the complex data curation required for sequential modeling in LVM~\cite{bai2024sequential}, 
pre-trained video generation models benefit from the inherent sequential structure of video data, allowing them to naturally capture temporal dependencies without the need for extensive task- and modality-specific annotations.
Inspired by LLaVA~\cite{liu2023visual}, which adapts a single generative backbone pre-trained on large-scale corpora to a variety of downstream tasks via supervised fine-tuning (SFT), we formulate the central hypothesis of this work:
\emph{Could a single large video-generation model, pre-trained once for synthesis, serve as a universal visual backbone that can be efficiently adapted to a broad range of vision tasks through SFT?}

To investigate this hypothesis, we propose \textbf{UniVid}, a unified paradigm that fine-tunes a pre-trained video diffusion transformer (DiT) to address \textit{generation and pixel-level understanding vision tasks} without any task-specific architectural modifications. 
Within this framework, both image and video tasks are naturally organized as \textit{visual sentences} that align with the temporal dimension of video data. 
As illustrated in the right block of~\figref{teaser}(a), each training sample is structured as $A \rightarrow A' \rightarrow B \rightarrow B'$, where the context $(A, A', B)$ defines the vision task and specifies the desired output modality.
We evaluate the generalization capacity of UniVid from two perspectives.
First, while LVM~\cite{bai2024sequential} is limited to uni-modal contexts (\ie either images or videos alone) at inference, we examine whether UniVid can accommodate \textbf{cross-modal contexts}---where the output modality is inferred from mixed image-video inputs. 
Second, we investigate UniVid’s capability for \textbf{cross-source tasks}, such as depth estimation from natural videos to annotated data, even without the multi-source pre-training required by LVM~\cite{bai2024sequential}.
As shown in~\figref{teaser}(b), despite being pre-trained solely on continuous natural video data, our model adapts effectively to both cross-modal and cross-source tasks through SFT. 
Importantly, under this unified paradigm, the distinction between generation and understanding tasks is reduced to the ordering of elements within the visual sentence.
These findings highlight the potential of pre-trained video generation models as a unified pre-training backbone for general-purpose visual modeling.

Our contributions are summarized as follows:
\begin{compactitem}
    \item To the best of our knowledge, we are the first to explore unified vision modeling using a pre-trained video generation model, eliminating the need for task- and modality-specific pre-training data.
    \item We propose \textbf{UniVid}, a unified framework that leverages lightweight SFT to efficiently adapt a pre-trained video DiT to a broad range of image and video tasks, without any task-specific architectural modifications.
    \item Extensive experiments show that UniVid effectively generalizes to both cross-modal and cross-source scenarios, highlighting its potential as a unified and scalable foundation for general-purpose vision modeling.
\end{compactitem}

\begin{table}[t] 
  \centering
  {%
  \renewcommand{\arraystretch}{0.9}
   \small
  \vspace{-1 mm}
  \begin{tabular}{>{\centering\arraybackslash}p{1.2cm}>{\centering\arraybackslash}p{1.2cm}>{\centering\arraybackslash}p{1.2cm}>{\centering\arraybackslash}p{1.2cm}>{\centering\arraybackslash}p{1.2cm}}
    \toprule
    \multirow{2}{*}{Context} & 
    \multicolumn{2}{c}{Example} & 
    \multicolumn{2}{c}{Query} \\
    \cmidrule(lr){2-3} \cmidrule(lr){4-5}
    & $A$ & $A'$ & $B$ & $B'$ \\
    \midrule
    \uppercase\expandafter{\romannumeral1} & Video & Video & Video & Video\\ 
    \uppercase\expandafter{\romannumeral2} & Image & Image & Image & Image \\ 
    \uppercase\expandafter{\romannumeral3} & Image & Image & Video & Video\\ 
    \uppercase\expandafter{\romannumeral4} & Image & Video &Image  &Video \\ 
    \bottomrule
  \end{tabular}
   
  }%
  \vspace{-3 mm}
   \caption{\textbf{Four Types of Visual Contexts.}}
  \label{tab:patterns}
  \vspace{-6 mm}
\end{table}

\section{Related Works}
\label{related}
\Paragraph{Large Vision Models.} Recent advancements in universal vision frameworks predominantly follow two paradigms: image-resembling generation~\cite{bar2022visual,wang2023images,wang2024lavin} and sequential modeling~\cite{sun2024x,bai2024sequential}.
Image-resembling generation methods~\cite{bar2022visual,wang2023images,wang2024lavin} reformulate diverse vision tasks as image inpainting problems, enabling models to make predictions through generating masked regions.
Sequential modeling~\cite{sun2024x,bai2024sequential} approaches draw inspiration from LLMs, treating visual data as sequences of discrete tokens and optimizing models via next-token prediction.
However, these approaches heavily rely on large-scale annotated data to construct task-specific training samples, which is resource-intensive and hinders scalability.
In contrast, we demonstrate that a model trained solely on continuous video data can be effectively adapted to a broad range of vision tasks.

\Paragraph{Vision In-Context Learning.} In-context learning, where models perform tasks prompted by examples, has been extensively studied in LLMs~\cite{brown2020language}.
Inspired by this success, many efforts~\cite{bar2022visual,bai2024sequential,wang2023images,sun2024x,wang2024lavin,huang2024context,tan2024ominicontrol,zhang2025easycontrol,chen2025edit,mao2025ace} have extended the paradigm to vision, demonstrating its potential across a wide range of vision tasks.
Early methods~\cite{bai2024sequential,wang2023images,bar2022visual,sun2024x} primarily rely on sequential models trained on large-scale annotated input-output pairs to perform vision tasks.
More recently, in-context generation capability has been demonstrated in DiT-based text-to-image (T2I) models~\cite{huang2024context,tan2024ominicontrol,zhang2025easycontrol,chen2025edit,mao2025ace} for controllable image generation and manipulation.
Similarly, the DiT-based video model captures temporal dependencies through full attention across frames, which we leverage to adapt the pre-trained model to a wide range of vision tasks.

\Paragraph{Video Generation Models.} Recent progress in video generation has been driven by both autoregressive~\cite{yan2021videogpt, hong2022cogvideo, villegas2022phenaki, kondratyuk2023videopoet, xie2024show, liu2024mardini} and diffusion-based~\cite{wang2023modelscope, chen2024videocrafter2, khachatryan2023text2video, qiu2025skyreels, lu2023vdt, ma2024latte, yang2024cogvideox, kong2024hunyuanvideo, wang2025wan} architectures.
Autoregressive models~\cite{yan2021videogpt, hong2022cogvideo, villegas2022phenaki, kondratyuk2023videopoet,xie2024show,liu2024mardini} compress video frames into discrete tokens and employ transformers to generate video sequences token by token.
Early diffusion-based video models~\cite{wang2023modelscope, chen2024videocrafter2, khachatryan2023text2video} extend U-Net~\cite{ronneberger2015u} architectures originally designed for T2I generation, lacking temporal consistency.
%
Recent hybrid architectures~\cite{qiu2025skyreels, lu2023vdt, ma2024latte, yang2024cogvideox, kong2024hunyuanvideo, wang2025wan} adopt DiT framework with advanced attention mechanism, yielding improved generation quality and temporal coherence.
In this work, we take the Wan model~\cite{wang2025wan} as a strong foundation to assess the effectiveness of video generative pre-training for downstream vision tasks.
\section{Methodology}
\label{sec:analysis}
In this section, we first review existing vision sequential models in~\subsecref{pre} and define the problems we explore in~\subsecref{prob}.  
Based on the discussion, we introduce our method, experimental setup, and results analysis in~\subsecref{method}.  
%

\begin{figure}[t]
    \vspace{-1 mm} %
    \centering
    \includegraphics[width=0.98\linewidth]{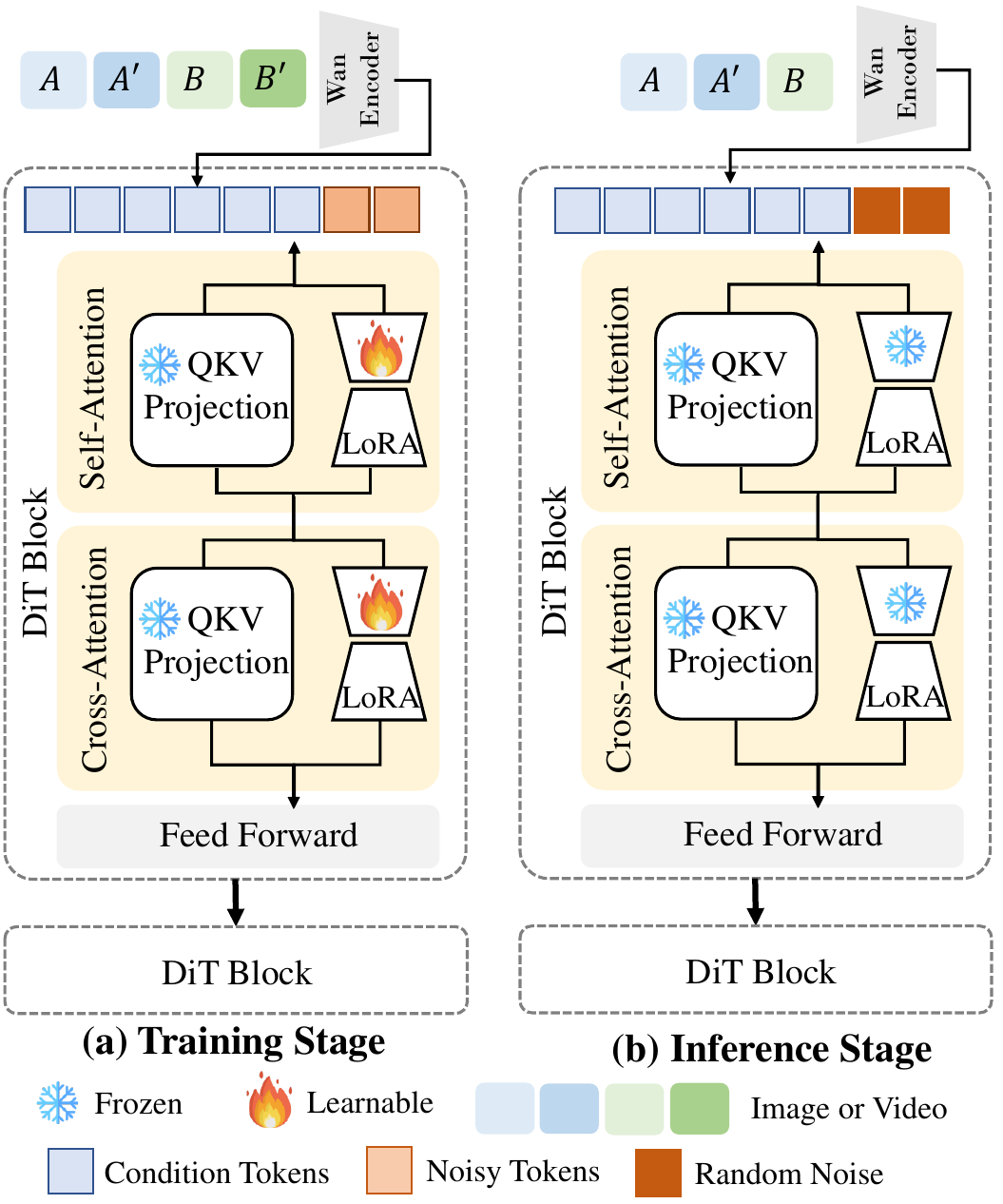}
    \vspace{-2 mm}
    \caption{\textbf{The framework of UniVid.}
    }
    \label{fig:framework}
    \vspace{-4 mm}
\end{figure}
\begin{figure*}[t!]
    \centering
\includegraphics[width=0.99\linewidth]{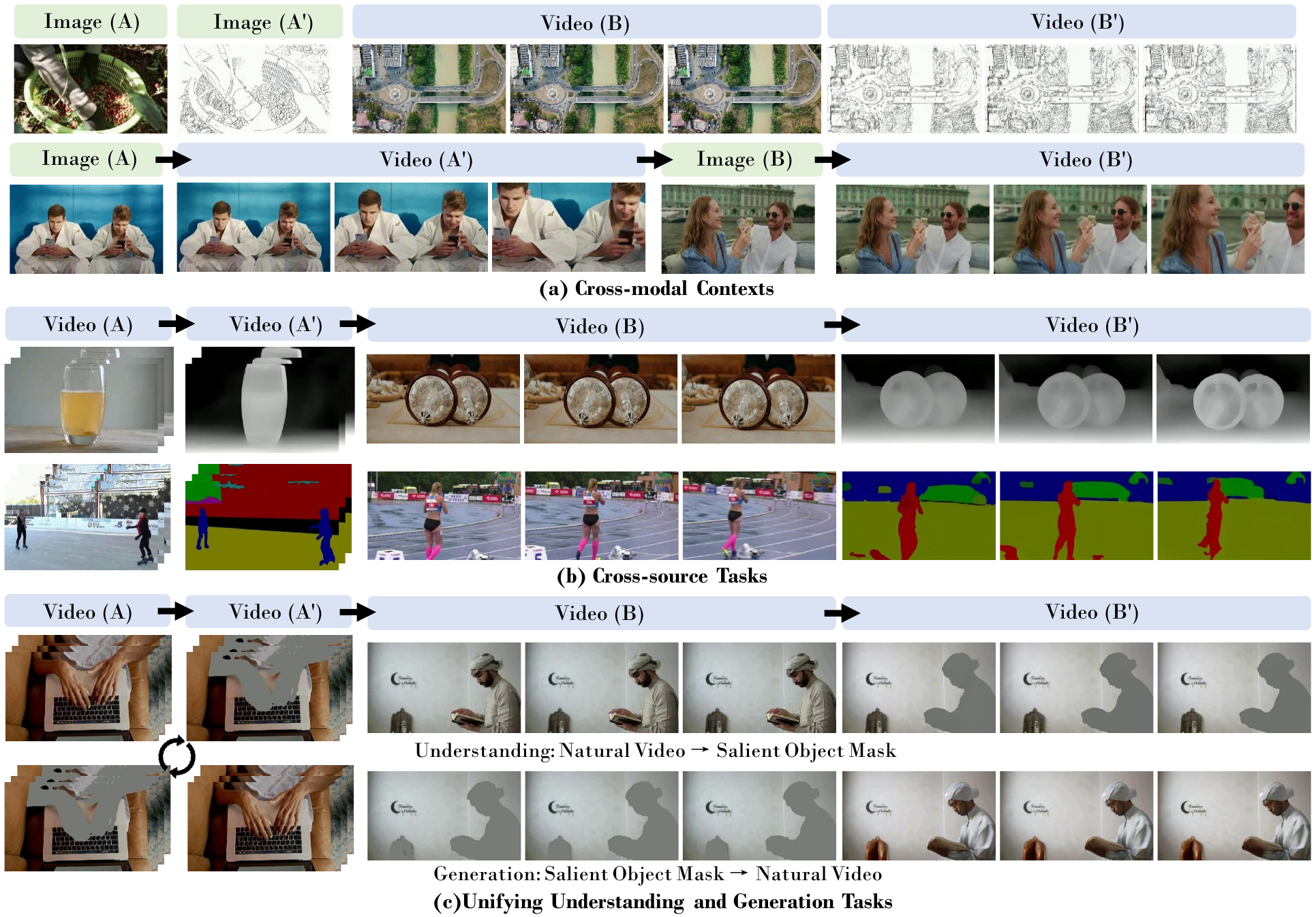}  
    \vspace{-2.5 mm}
  \caption{\textbf{Main observations.}
The top colored row serves as a legend indicating the modality and role of each clip shown below.
The following figure follows the same format.
(a) The model infers the correct output modality from cross-modal contexts.  
(b) Despite being pre-trained solely on natural video data, it generalizes to cross-source understanding tasks.  
(c) Under the UniVid framework, understanding and generation tasks are unified and can be converted by reordering the visual sentence.
    }
    \vspace{-4 mm}
    \label{fig:tasks}
\end{figure*}
\begin{figure*}[t!]
\centering
\includegraphics[width=0.99\linewidth]{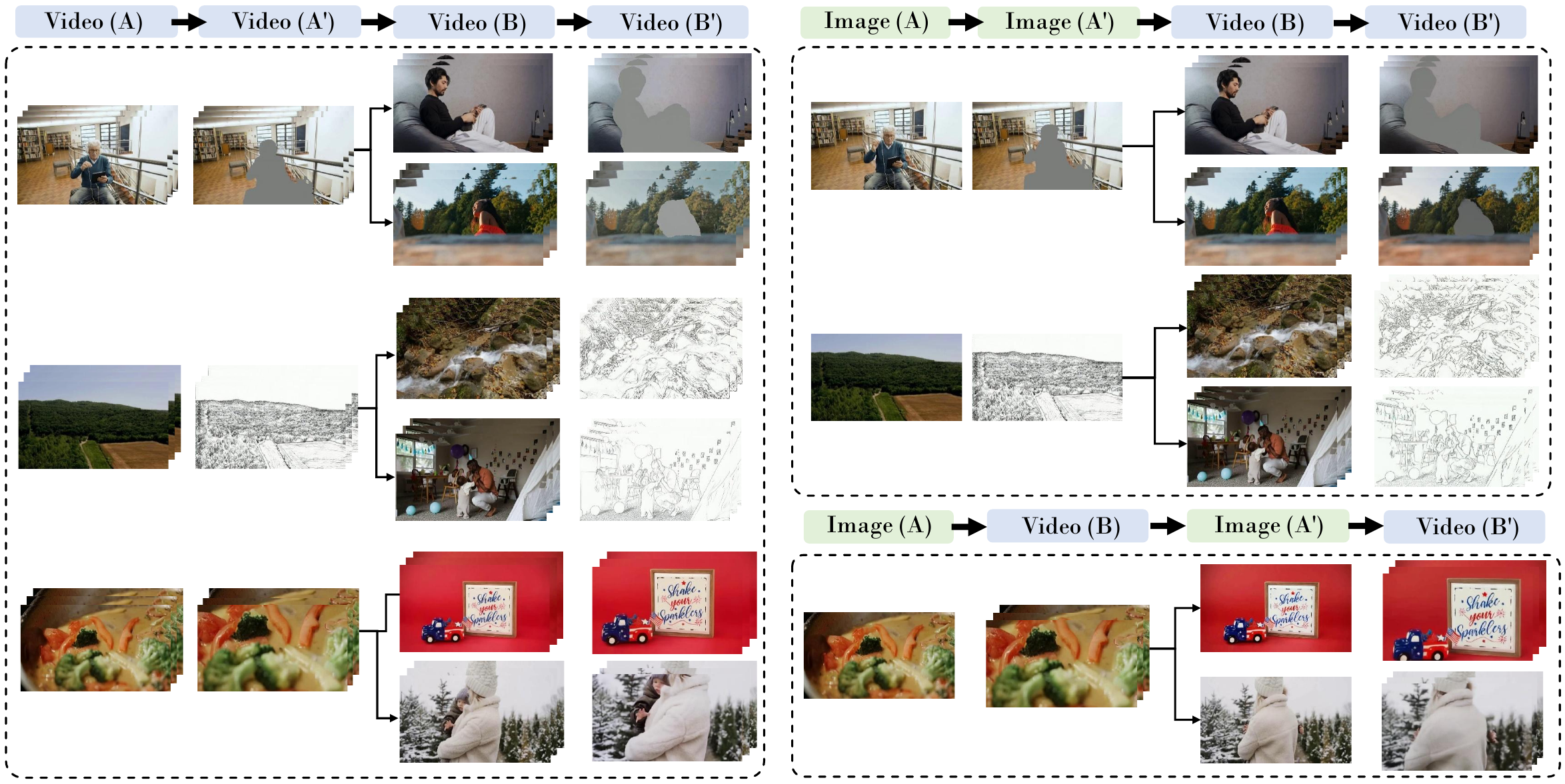}  
    \vspace{-3 mm}
    \caption{\textbf{Performance across diverse vision tasks and context formats.}
We show results for scribble map transfer, motion transfer, and salient object tracking under various visual contexts.
Each task is fine-tuned independently within each context configuration, demonstrating that the pre-trained video generation model adapts well across all applicable settings listed in~\tabref{patterns}.
With a fixed example pair, outputs change with the query, reflecting context-based inference.
    }
    \vspace{-5 mm}
    \label{fig:additional}
\end{figure*}

\subsection{Preliminaries: Limitations of Sequential Vision Models}

\label{subsec:pre}
Recent advances in sequential vision modeling, exemplified by the LVM~\cite{bai2024sequential}, aim to unify diverse vision tasks under a single framework. However, LVM~\cite{bai2024sequential} faces two significant limitations:

\Paragraph{(a) Reliance on Annotated Pairs for Pre-training.} 
As shown in~\figref{teaser}(a), LVM~\cite{bai2024sequential} requires large-scale, task- and modality-specific annotated pairs for pre-training in order to support a wide range of downstream tasks. 
This data curation process is labor-intensive and fundamentally restricts scalability. 
In this work, we question whether annotated pairs are truly necessary at the pre-training stage.

\Paragraph{(b) Limited Generalization Beyond Single-Modality Contexts.} 
As shown in~\figref{teaser}(b), despite being pre-trained on heterogeneous data, LVM~\cite{bai2024sequential} is limited to tasks within image-only or video-only contexts at inference, leaving cross-modal scenarios largely underexplored. 
Furthermore, LVM~\cite{bai2024sequential} is restricted to video extrapolation and fails to generalize to cross-source video tasks, such as semantic segmentation, despite relevant annotations in pre-training.

\textit{These limitations motivate us to seek an alternative unified vision modeling paradigm that reduces data annotation burdens and extends generalization across modalities and sources.}

%

\subsection{Problem Formulation: Unified Visual Sentence Paradigm}
\label{subsec:prob}

To address these challenges, we revisit the video generation model, treating it as a naturally pre-trained visual sequential learner.
We reformulate a broad set of vision tasks using a unified \textit{visual sentence} paradigm: given an example input-output pair $A \rightarrow A'$, the model is tasked to predict $B'$ for a query $B$, with $(A, A', B)$ collectively defining the task context and output modality.
We focus on evaluating the adaptation capacity of a pre-trained video generation model from two key perspectives:(a) \textbf{Cross-modal generalization}: the ability to handle tasks prompted by mixed-modal contexts (\eg combining images and videos), as illustrated in~\tabref{patterns} \uppercase\expandafter{\romannumeral3} and \uppercase\expandafter{\romannumeral4}.
(b) \textbf{Cross-source generalization}: the ability to perform vision tasks across heterogeneous data sources, spanning both \textit{pixel-level understanding and generation tasks}.
Our central question is: \textbf{\emph{Can a video generation model pre-trained solely on natural videos adapt to diverse vision tasks within a unified visual sentence paradigm, even when the contexts span modalities and data sources?}}

%

\begin{figure*}[t!]
    \centering
\includegraphics[width=0.99\linewidth]{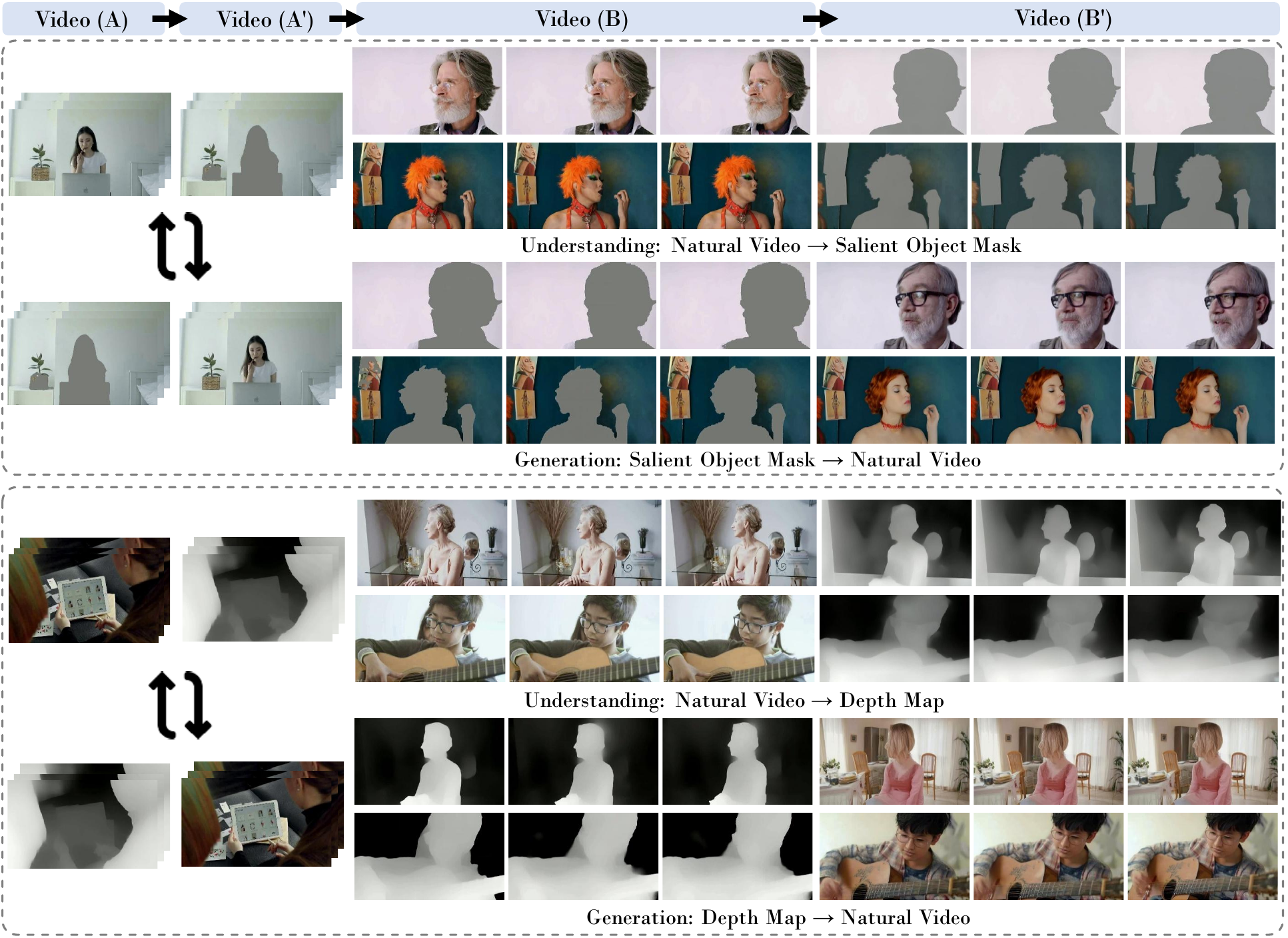}  
    \vspace{-3 mm}
  \caption{\textbf{Unified understanding and generation tasks.}
    Our proposed UniVid allows flexible switching between understanding and generation tasks by simply reordering visual sentences.
    }
    \vspace{-5 mm}
    \label{fig:generate}
\end{figure*}

\subsection{Video Generative Pre-training with SFT for Unified Vision Modeling}
\label{subsec:method}

We start from a video DiT model \ie Wan~\cite{wang2025wan}, pre-trained exclusively on continuous natural video data, as our unified backbone. 
To enable adaptation to diverse vision tasks, we employ Low-Rank Adaptation (LoRA) modules for efficient SFT.

\Paragraph{Data Structure.} 
We leverage the temporal dimension of video by representing each input-output pair as a sequence of video clips concatenated along the time axis. 
As shown in~\figref{framework}(a), each training sample is structured as a visual sentence $V = [A, A', B, B']$.
$A$ and $A'$ comprise an example pair that demonstrates a reference vision task $A \rightarrow A'$ (\eg a source video and its scribble map).
$B$ is the query input, and $B'$ is the expected output, which should undergo the same transformation as $A \rightarrow A'$.
Each element can be either an image or a video segment, allowing unified processing of diverse contexts.

\Paragraph{Fine-Tuning and Inference.} 
During training, $V$ is first embedded into latent tokens.
We treat the tokens corresponding to $(A, A', B)$ as the context $C$, which remains clean during training.
Noise is added only to the target clip $B'$, producing noisy latent tokens $z_t$, as shown in~\figref{framework}(a).
The complete token sequence is then processed by the DiT blocks using 3D full attention~\cite{wang2025wan}, enabling cross-clip interactions throughout the temporal dimension.
We insert LoRA modules into both Cross-Attention (CA) and Self-Attention (SA) layers for efficient adaptation.
%
%
At inference, the model generates $B'$ conditioned on $(A, A', B)$, with clean context tokens guiding the generation process as illustrated in~\figref{framework}(b).

\begin{figure*}[t!]
    \centering
\includegraphics[width=0.99\linewidth]{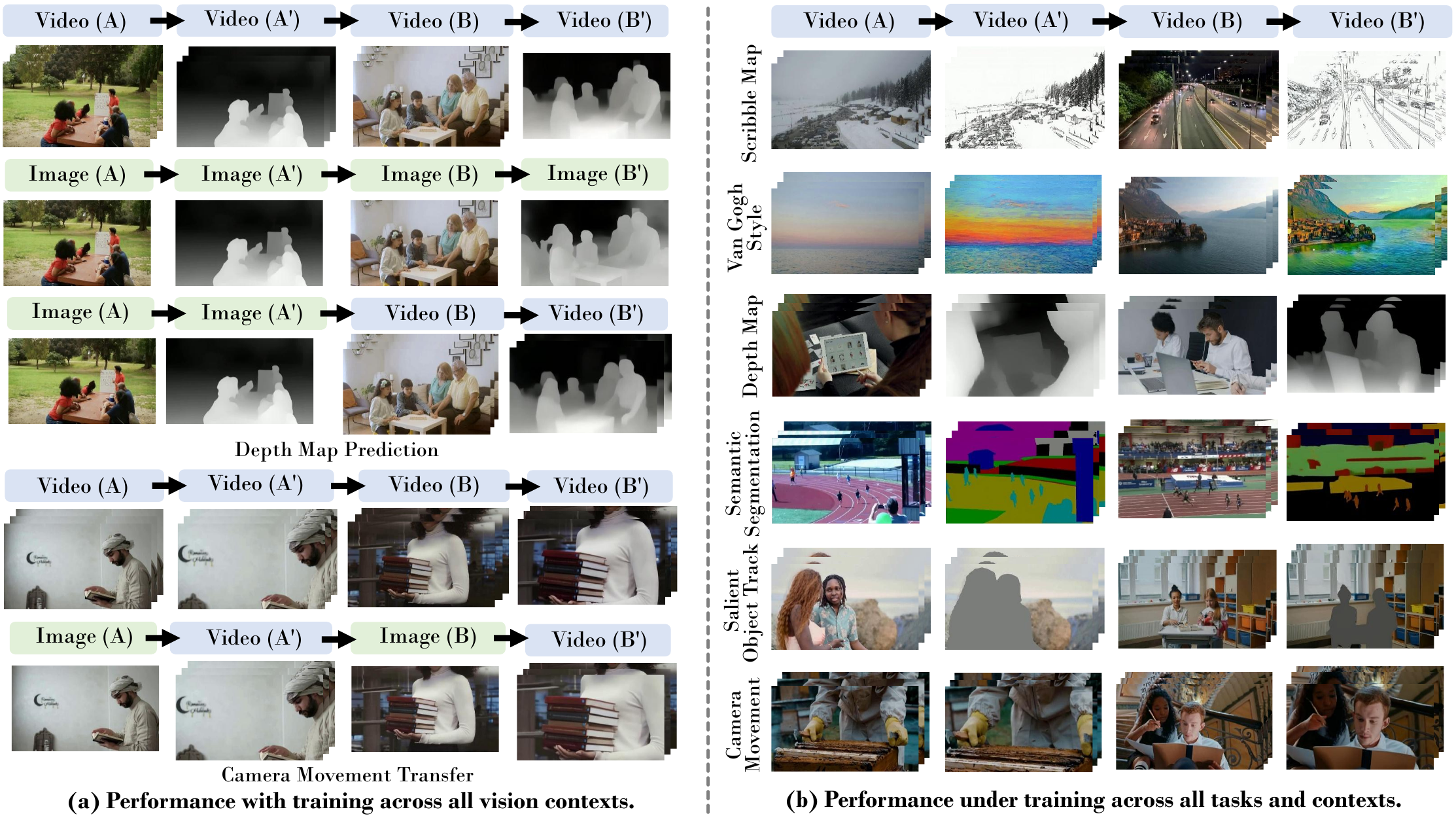}    
\vspace{-2.5 mm}
  \caption{\textbf{Results of mixed fine-tuning strategy.}
    (a) When fine-tuned on visual sequences spanning all context types, the model can dynamically infer the output modality at inference time based on the context $(A, A', B)$.
    (b) We co-train the model on all vision tasks, each covering all applicable contexts.  
    Under this mixed setup, the model consistently performs well across all tasks and contexts, demonstrating strong generalization.  
    Additional results are shown in the appendix.
    }
    \vspace{-4 mm}
        \label{fig:mix}
\end{figure*}

\Paragraph{Experimental Protocol.}
\label{subsec:setup}
To systematically assess adaptation and generalization, we collect paired data across six representative tasks, including generation tasks (1) scribble map transfer, (2) Van Gogh style transfer and (3) camera movement transfer and understanding tasks (4) depth map prediction, (5) semantic segmentation prediction, and (6) salient object tracking.
Each task consists of only $20$ training samples.
To validate the question posed in~\subsecref{prob}, we conduct straightforward fine-tuning experiments: 
First, for each vision task, we fine-tune the video generation model on its applicable contexts listed in~\tabref{patterns} separately.
Additionally, we construct generation variants of tasks (4) and (6) by reversing them into conditional generation settings to further establish its generalization.

\Paragraph{Results and Analysis.}
Based on our experimental results (see~\figref{tasks}), we summarize the following three key observations regarding the model’s adaptability to diverse vision tasks:

\textbf{Observation 1:} \emph{Robust cross-modal adaptation.}
Although pre-trained solely on continuous video data, the fine-tuned model effectively interprets manually composed visual sentences and flexibly handles various task contexts, including those spanning multiple modalities (see~\figref{tasks}(a)).

\textbf{Observation 2:} \emph{Effective cross-source generalization.}
Despite only being exposed to natural video data during pre-training, the model adapts successfully to cross-source tasks such as predicting depth maps from natural video data (see~\figref{tasks}(b)).

\textbf{Observation 3:} \emph{Unified formulation of understanding and generation tasks.}
As illustrated in~\figref{tasks}(c), the model can perform both understanding and generation tasks by simply reordering the elements within a visual sentence. 
This demonstrates that, under the unified paradigm, these tasks are seamlessly interchangeable through sentence organization.

\section{Experiments}
In this section, we first present the implementation details in~\subsecref{implementation}.
In~\subsecref{generalization}, we assess how well UniVid generalizes to a range of vision tasks under varied visual contexts.
Next, we explore the results of mixed fine-tuning strategies in~\subsecref{finetuning} and the effects of shot number in~\subsecref{shot}.
Quantitative comparisons with LVM~\cite{bai2024sequential} are provided in~\subsecref{quantitative} to validate the effectiveness of our approach. 
Finally, we discuss the remaining limitations and future work in~\subsecref{discussion}.


\subsection{Implementation Details.}
\label{subsec:implementation}
We employ Wan2.1-T2V-1.3B as the backbone video generation model for our experiments.
The number of frames in each clip of $(A, A', B, B')$ is set to $1$ for image modality and $17$ for video modality.
We use the Wan Encoder~\cite{wang2025wan} to embed the four clips separately.
During fine-tuning, we set the LoRA rank to 16, the learning rate to $1 \times 10^{-4}$, and the batch size to 1.
%
Please refer to appendix for additional details.

\subsection{Generalization across Tasks and Contexts}
\label{subsec:generalization}
\Paragraph{Task-level Generalization.}  
We evaluate UniVid on the six tasks under diverse context configurations (\tabref{patterns}).
Specifically, the camera motion transfer task is evaluated under contexts \uppercase\expandafter{\romannumeral1} and \uppercase\expandafter{\romannumeral4}, while the other tasks are trained under contexts \uppercase\expandafter{\romannumeral1}, \uppercase\expandafter{\romannumeral2}, and \uppercase\expandafter{\romannumeral3}.  
As shown in~\figref{additional}, UniVid demonstrates strong adaptability across modalities and data sources.
%
%
Please refer to appendix for remaining results.


\Paragraph{Unified Understanding and Generation.}  
We reverse the visual sentence structure from $(A \rightarrow A' \rightarrow B \rightarrow B')$ to $(A' \rightarrow A \rightarrow B' \rightarrow B)$, converting understanding tasks into generation tasks.  
As illustrated in~\figref{generate}, the fine-tuned model generates coherent videos under both sequence orderings, highlighting the unified formulation of both understanding and generation.
%
%

\Paragraph{Context-Conditioned Inference.}  
As shown in~\figref{additional}, given a fixed example pair $(A, A')$, the output $B'$ changes based on query $B$, confirming the model's context-conditioned reasoning capability.

\subsection{Mixed Fine-tuning Strategy}
\label{subsec:finetuning}

Building on the results showing that the video generation model performs well when each vision task is fine-tuned with a single visual context, as demonstrated in~\figref{tasks} and~\figref{additional}, we further explore its adaptability under joint training regimes.
We consider two configurations: (1) training each task with a mixture of visual contexts, and (2) jointly training all tasks, where each task is exposed to all applicable contexts.

\Paragraph{Per-task Mixed Context Fine-tuning.}
%
For the camera motion transfer task, the training data is composed of a mixture of contexts \uppercase\expandafter{\romannumeral1} and \uppercase\expandafter{\romannumeral4}, while the other tasks are trained under contexts \uppercase\expandafter{\romannumeral1}, \uppercase\expandafter{\romannumeral2}, and \uppercase\expandafter{\romannumeral3}.
As shown in~\figref{mix}(a), when trained on mixed contexts, the model can automatically adapt to the specific context $(A, A', B)$ and produce consistent results in correct modality, demonstrating strong generalization within the unified training paradigm.
Please refer to appendix for the results of other tasks.
%

\Paragraph{Joint Multi-task Fine-tuning.}
We co-train the model on all six vision tasks, with each task provided $20$ training examples covering all applicable contexts.
%
%
\figref{mix} presents results for the six vision tasks under Context \uppercase\expandafter{\romannumeral1}, with the remaining three contexts provided in appendix.
These results demonstrate that even with limited fine-tuning data mixed from multiple vision tasks and contexts, the model generalizes robustly across diverse vision tasks and contexts, validating its potential as a unified foundation for visual modeling.
We further report quantitative results of joint multi-task fine-tuning, employing CLIP-T for style transfer task.
For the camera movement task, we extract the first, middle, and last frames and ask GPT-4o to rate the quality on a scale from $0$ to $10$.
Remaining tasks are measured by pixel-space RMSE.
As shown in~\tabref{training_comparison}, the co-training strategy consistently achieves better performance than separate fine-tuning across all tasks.

\subsection{Impact of Shot Number}
\label{subsec:shot}
We investigate the effect of shot number by fine-tuning three separate models using 4-shot, 6-shot, and 8-shot configurations, respectively. 
Each model is then evaluated under all three shot settings.
Specifically, the 4-shot, 6-shot, and 8-shot settings are defined as follows:

\begin{compactitem}
    \item \textbf{$4$ shots}: $A$ $\rightarrow$ $A'$ $\rightarrow$ $B$ $\rightarrow$ $B'$
    \item \textbf{$6$ shots}: $A$ $\rightarrow$ $A'$ $\rightarrow$ $B$ $\rightarrow$ $B'$ $\rightarrow$ $C$ $\rightarrow$ $C'$
    \item \textbf{$8$ shots}: $A$ $\rightarrow$ $A'$ $\rightarrow$ $B$ $\rightarrow$ $B'$ $\rightarrow$ $C$ $\rightarrow$ $C'$ $\rightarrow$ $D$ $\rightarrow$ $D'$
\end{compactitem}

In all cases, the final clip (e.g., $B'$, $C'$, or $D'$) is the target to be generated. 
We evaluate on a classic understanding task (depth estimation) and a generation task (style transfer) on uni-image Context \uppercase\expandafter{\romannumeral2}, reporting results in~\tabref{shot_number}.
%
%
For depth estimation, performance tends to degrade when the number of test shots exceeds that used in fine-tuning, as the model never saw depth maps during pre-training and relies solely on fine-tuning supervision.
In contrast, style transfer remains stable across shot counts since similar visual data were seen during pre-training.
%
%
%
%
While longer contexts yield better results, they also increase inference time. For consistent analysis, we adopt a four-shot setting in this paper.
\begin{table}[t]
\centering
\footnotesize
\begin{tabular}{>{\centering\arraybackslash}p{4.3cm}>{\centering\arraybackslash}p{1.2cm}>{\centering\arraybackslash}p{1.5cm}}
\toprule
\multirow{2}{*}{\parbox{2.5cm}{\centering Task}} & \multicolumn{2}{c}{Training Strategy} \\
\cmidrule(lr){2-3}
& Separate & Co-training \\
\midrule
\parbox{4.0cm}{\centering Style  Transfer (CLIP-T $\uparrow$)} & 18.33 & \textbf{24.01} \\
\midrule
\parbox{4.0cm}{\centering Camera  Movement (GPT-4o $\uparrow$)} & 6.33 & \textbf{6.73} \\
\midrule
\parbox{4.0cm}{\centering Scribble Map  (RMSE $\downarrow$)} & 61.03 & \textbf{51.94} \\
\midrule
\parbox{4.0cm}{\centering Depth  Estimation  (RMSE $\downarrow$)} & 74.16 & \textbf{2.55} \\
\midrule
\parbox{4.3cm}{\centering Semantic  Segmentation  (RMSE $\downarrow$)} & 126.12 & \textbf{123.03} \\
\midrule
\parbox{4.0cm}{\centering Salient Object  Track (RMSE $\downarrow$)} & 33.35 & \textbf{30.59} \\
\bottomrule
\end{tabular}
\vspace{-2 mm}
\caption{\textbf{Quantitative comparisons between different training strategies.}}
\label{tab:training_comparison}
\end{table}
\begin{table}[t]
\centering
\small
\begin{tabular}{
>{\centering\arraybackslash}p{0.5cm} 
>{\centering\arraybackslash}p{0.5cm} 
>{\centering\arraybackslash}p{1.1cm} 
>{\centering\arraybackslash}p{2.0cm} 
>{\centering\arraybackslash}p{2.0cm} 
}
\toprule
\parbox{0.5cm}{\centering FT \\shots} & \parbox{0.5cm}{\centering Test \\shots} & Time (s) & \parbox{2.3cm}{\centering Depth Estimation \\ (RMSE $\downarrow$)} & \parbox{1.8cm}{\centering Style Transfer \\ (CLIP-T $\uparrow$)} \\
\midrule     
\multirow{3}{*}{4} & 4 &13.37 & 61.03 & 21.28 \\
& 6 & 19.36& 74.25 & 21.00 \\
& 8 & 25.47& 86.58 & 20.93 \\
\midrule
\multirow{3}{*}{6} & 4 &13.37 &\textbf{44.85} & 21.16 \\
& 6 &19.36 & 54.51 & 21.00 \\
& 8 &25.47 & 47.84 & 21.01 \\
\midrule
\multirow{3}{*}{8} & 4 &13.37 & 49.5 & \textbf{21.65} \\
& 6 &19.36 & 52.07 & 21.57 \\
& 8 & 25.47& 55.16 & 21.51 \\
\bottomrule
\end{tabular}
\vspace{-2 mm}
\caption{\textbf{Effects of the shot number.}}
\label{tab:shot_number}
\vspace{-8 mm}
\end{table}

\subsection{Quantitative Comparisons}
\label{subsec:quantitative}

To assess the effectiveness of \textbf{UniVid}, we compare it with LVM~\cite{bai2024sequential} across five tasks:
%
%
\begin{compactitem}
    \item  \textbf{Van Gogh style transfer} is evaluated with CLIP-T (alignment with ``Van Gogh style'') and CLIP-I (consistency with reference image $A'$).
    \item \textbf{Edge map prediction} on the BIPED~\cite{poma2020dense} dataset, is evaluated by fixed contour threshold (ODS), per-image best threshold (OIS), and average precision (AP).
    \item \textbf{Semantic segmentation} is conducted on the ADE20K dataset~\cite{zhou2019semantic}, measured by mean Intersection over Union (mIoU) and pixel accuracy (pAcc).
    \item \textbf{Depth estimation} is evaluated on the NYU-v2 dataset~\cite{silberman2012indoor}, reporting the percentage of pixels within various $\delta$ thresholds, absolute relative error (AbsRel), squared relative error (SqRel), root mean square logarithmic error (RMSELog), and scale-invariant logarithmic error (SILog).
    \item \textbf{Surface normal estimation} is also evaluated on the NYU-v2 dataset~\cite{silberman2012indoor}, assessed by Mean Angular errors (Mean) and Median Angular Errors (Med), along with accuracy under thresholds of $5^\circ$, $11.25^\circ$, and $30^\circ$.
\end{compactitem}
%
%
All tasks are trained on small subsets of the standard training sets (such as 65 of 175K in NYU-v2~\cite{silberman2012indoor}) and evaluated on the full test splits, whereas LVM~\cite{bai2024sequential} uses the full training split.
%
As shown in~\tabref{combined_performance}, \tabref{depth} and \tabref{normal}, despite being trained on limited data, our method consistently outperforms LVM~\cite{bai2024sequential}, demonstrating strong generalization capability to both visual generation and understanding tasks.


\begin{table}[!t]
\centering
\footnotesize 
\renewcommand{\arraystretch}{1.1}
\begin{tabular}{>{\arraybackslash}p{0.5cm}>{\centering\arraybackslash}p{1.03cm}>{\centering\arraybackslash}p{0.95cm}>{\centering\arraybackslash}p{0.4cm}>{\centering\arraybackslash}p{0.4cm}>{\centering\arraybackslash}p{0.4cm}>{\centering\arraybackslash}p{0.6cm}>{\centering\arraybackslash}p{0.6cm}}
\toprule
\multirow{2}{*}{Method} & \multicolumn{2}{c}{\parbox{0.8cm}{\centering Style \\ Transfer}} & \multicolumn{3}{c}{\parbox{2.0cm}{\centering Edge Map\\ Prediction}} & \multicolumn{2}{c}{\parbox{1.2cm}{\centering Semantic \\ Segmentation}} \\
\cmidrule(lr){2-3} \cmidrule(lr){4-6} \cmidrule(l){7-8}
& CLIP-T$\uparrow$ & CLIP-I$\downarrow$ & ODS$\uparrow$ & OIS$\uparrow$ & AP$\uparrow$ & mIoU$\uparrow$ & pACC$\uparrow$ \\
\midrule
LVM & 16.24 & 0.712 & 0.656 & 0.678 & 0.630 & 1.423 & 23.30 \\
Ours & \textbf{19.76} & \textbf{0.670} & \textbf{0.873} & \textbf{0.877} & \textbf{0.871} & \textbf{8.712} & \textbf{53.13} \\
\bottomrule
\end{tabular}
\vspace{-2.5 mm}
\caption{\textbf{Comparison across style transfer, edge map prediction, and semantic segmentation.}}
\label{tab:combined_performance}
\vspace{-2 mm}
\end{table}
\begin{table}[!t]
\centering
\footnotesize
\begin{tabular}{>{\centering\arraybackslash}p{0.6cm}>{\centering\arraybackslash}p{0.3cm}>{\centering\arraybackslash}p{0.3cm}>{\centering\arraybackslash}p{0.3cm}>{\centering\arraybackslash}p{0.8cm}>{\centering\arraybackslash}p{0.6cm}>{\centering\arraybackslash}p{1.0cm}>{\centering\arraybackslash}p{0.8cm}}
\toprule
Method & $\delta_1$$\uparrow$ & $\delta_3$$\uparrow$ & $\delta_3$$\uparrow$ & AbsRel$\downarrow$ & SqRel$\downarrow$ & RMSElog$\downarrow$ & SILog$\downarrow$ \\
\midrule
LVM & 0.15 & 0.31 & 0.48 & 0.53 & \textbf{0.18} & 1.15 & 72.91 \\
Ours & \textbf{0.43} & \textbf{0.76} & \textbf{0.91} & \textbf{0.27} & 0.28 & \textbf{0.42} & \textbf{30.74}  \\
\bottomrule
\end{tabular}
\vspace{-3 mm}
\caption{\textbf{Depth estimation performance.} }
\label{tab:depth}
\vspace{-2 mm}
\end{table}
\begin{table}[!t]
\footnotesize
\centering
\begin{tabular}{>{\centering\arraybackslash}p{0.6cm} >{\centering\arraybackslash}p{0.85cm} >{\centering\arraybackslash}p{0.85cm} >{\centering\arraybackslash}p{1.1cm}>{\centering\arraybackslash} p{1.3cm}>{\centering\arraybackslash} p{1.0cm}}
\toprule
Method & Mean $\downarrow$ & Med $\downarrow$ & $5^\circ$$\uparrow$ & $11.25^\circ$$\uparrow$ & $30^\circ$$\uparrow$ \\
\midrule
LVM  & 30.76 & 13.73 & 24.70\% & 45.10\% & 65.98\% \\
Ours  & \textbf{29.84} & \textbf{13.52} & \textbf{25.22\%} & \textbf{45.57\%} & \textbf{67.53\%} \\
\bottomrule
\end{tabular}
\vspace{-3 mm}
\caption{\textbf{Surface normal estimation performance.}}
\label{tab:normal}
\vspace{-2 mm}
\end{table}
\subsection{Limitations and Future Work}
\label{subsec:discussion}
While our study validates a promising approach to unifying vision tasks using a video generation model, certain limitations remain.
%
The context length of Wan model~\cite{wang2025wan} is limited to $81$ frames per sequence, restricting the duration of each visual clip. 
Additionally, due to the inherent randomness of generative processes, label consistency across instance types in the segmentation task cannot be guaranteed.
%
In the future, we plan to explore long-context video generation~\cite{gu2025long} architectures and mitigate ambiguity in understanding tasks such as segmentation.



\section{Conclusions}
In this work, we explore a new direction for building a unified visual backbone by reusing a pre-trained video generation model for various vision tasks through lightweight supervised fine-tuning.
Unlike prior approaches that rely on large-scale, task-specific data to train a visual sequential model, we simply fine-tune a pre-trained video generation model using minimal supervised data.
The visual data are organized into visual sentences, where the context defines both the vision task and the expected output modality.
%
%
Despite being pre-trained solely on natural, continuous videos without annotations, the fine-tuned model generalizes well to cross-modal and cross-source contexts.
Under this unified paradigm, understanding and generation tasks are differentiated only by the order of visual sentences and can be seamlessly interchanged.
Moreover, results show that with minimal supervision, the model can jointly adapt to diverse vision tasks and context types through single fine-tuning.
\clearpage
{
    \small
    \bibliographystyle{ieeenat_fullname}
    \bibliography{main}

\begin{thebibliography}{37}
\providecommand{\natexlab}[1]{#1}
\providecommand{\url}[1]{\texttt{#1}}
\expandafter\ifx\csname urlstyle\endcsname\relax
  \providecommand{\doi}[1]{doi: #1}\else
  \providecommand{\doi}{doi: \begingroup \urlstyle{rm}\Url}\fi

\bibitem[Bai et~al.(2024)Bai, Geng, Mangalam, Bar, Yuille, Darrell, Malik, and Efros]{bai2024sequential}
Yutong Bai, Xinyang Geng, Karttikeya Mangalam, Amir Bar, Alan~L Yuille, Trevor Darrell, Jitendra Malik, and Alexei~A Efros.
\newblock Sequential modeling enables scalable learning for large vision models.
\newblock In \emph{Proceedings of the IEEE/CVF Conference on Computer Vision and Pattern Recognition}, pages 22861--22872, 2024.

\bibitem[Bar et~al.(2022)Bar, Gandelsman, Darrell, Globerson, and Efros]{bar2022visual}
Amir Bar, Yossi Gandelsman, Trevor Darrell, Amir Globerson, and Alexei Efros.
\newblock Visual prompting via image inpainting.
\newblock \emph{Advances in Neural Information Processing Systems}, 35:\penalty0 25005--25017, 2022.

\bibitem[Chen et~al.(2024)Chen, Zhang, Cun, Xia, Wang, Weng, and Shan]{chen2024videocrafter2}
Haoxin Chen, Yong Zhang, Xiaodong Cun, Menghan Xia, Xintao Wang, Chao Weng, and Ying Shan.
\newblock Videocrafter2: Overcoming data limitations for high-quality video diffusion models.
\newblock In \emph{Proceedings of the IEEE/CVF Conference on Computer Vision and Pattern Recognition}, pages 7310--7320, 2024.

\bibitem[Chen et~al.(2025)Chen, Mao, Gu, and Shou]{chen2025edit}
Lan Chen, Qi Mao, Yuchao Gu, and Mike~Zheng Shou.
\newblock Edit transfer: Learning image editing via vision in-context relations.
\newblock \emph{arXiv preprint arXiv:2503.13327}, 2025.

\bibitem[Gu et~al.(2025)Gu, Mao, and Shou]{gu2025long}
Yuchao Gu, weijia Mao, and Mike~Zheng Shou.
\newblock Long-context autoregressive video modeling with next-frame prediction.
\newblock \emph{arXiv preprint arXiv:2503.19325}, 2025.

\bibitem[Guo et~al.(2025)Guo, Yang, Zhang, Song, Zhang, Xu, Zhu, Ma, Wang, Bi, et~al.]{guo2025deepseek}
Daya Guo, Dejian Yang, Haowei Zhang, Junxiao Song, Ruoyu Zhang, Runxin Xu, Qihao Zhu, Shirong Ma, Peiyi Wang, Xiao Bi, et~al.
\newblock Deepseek-r1: Incentivizing reasoning capability in llms via reinforcement learning.
\newblock \emph{arXiv preprint arXiv:2501.12948}, 2025.

\bibitem[Hong et~al.(2022)Hong, Ding, Zheng, Liu, and Tang]{hong2022cogvideo}
Wenyi Hong, Ming Ding, Wendi Zheng, Xinghan Liu, and Jie Tang.
\newblock Cogvideo: Large-scale pretraining for text-to-video generation via transformers.
\newblock \emph{arXiv preprint arXiv:2205.15868}, 2022.

\bibitem[Huang et~al.(2024)Huang, Wang, Wu, Shi, Dou, Liang, Feng, Liu, and Zhou]{huang2024context}
Lianghua Huang, Wei Wang, Zhi-Fan Wu, Yupeng Shi, Huanzhang Dou, Chen Liang, Yutong Feng, Yu Liu, and Jingren Zhou.
\newblock In-context lora for diffusion transformers.
\newblock \emph{arXiv preprint arXiv:2410.23775}, 2024.

\bibitem[Jiang et~al.(2025)Jiang, Han, Mao, Zhang, Pan, and Liu]{vace}
Zeyinzi Jiang, Zhen Han, Chaojie Mao, Jingfeng Zhang, Yulin Pan, and Yu Liu.
\newblock Vace: All-in-one video creation and editing.
\newblock \emph{arXiv preprint arXiv:2503.07598}, 2025.

\bibitem[Khachatryan et~al.(2023)Khachatryan, Movsisyan, Tadevosyan, Henschel, Wang, Navasardyan, and Shi]{khachatryan2023text2video}
Levon Khachatryan, Andranik Movsisyan, Vahram Tadevosyan, Roberto Henschel, Zhangyang Wang, Shant Navasardyan, and Humphrey Shi.
\newblock Text2video-zero: Text-to-image diffusion models are zero-shot video generators.
\newblock In \emph{Proceedings of the IEEE/CVF International Conference on Computer Vision}, pages 15954--15964, 2023.

\bibitem[Kondratyuk et~al.(2023)Kondratyuk, Yu, Gu, Lezama, Huang, Schindler, Hornung, Birodkar, Yan, Chiu, et~al.]{kondratyuk2023videopoet}
Dan Kondratyuk, Lijun Yu, Xiuye Gu, Jos{\'e} Lezama, Jonathan Huang, Grant Schindler, Rachel Hornung, Vighnesh Birodkar, Jimmy Yan, Ming-Chang Chiu, et~al.
\newblock Videopoet: A large language model for zero-shot video generation.
\newblock \emph{arXiv preprint arXiv:2312.14125}, 2023.

\bibitem[Kong et~al.(2024)Kong, Tian, Zhang, Min, Dai, Zhou, Xiong, Li, Wu, Zhang, et~al.]{kong2024hunyuanvideo}
Weijie Kong, Qi Tian, Zijian Zhang, Rox Min, Zuozhuo Dai, Jin Zhou, Jiangfeng Xiong, Xin Li, Bo Wu, Jianwei Zhang, et~al.
\newblock Hunyuanvideo: A systematic framework for large video generative models.
\newblock \emph{arXiv preprint arXiv:2412.03603}, 2024.

\bibitem[Liu et~al.(2023)Liu, Li, Wu, and Lee]{liu2023visual}
Haotian Liu, Chunyuan Li, Qingyang Wu, and Yong~Jae Lee.
\newblock Visual instruction tuning.
\newblock \emph{Advances in neural information processing systems}, 36:\penalty0 34892--34916, 2023.

\bibitem[Liu et~al.(2024)Liu, Liu, Zhou, Xu, Xie, Han, P{\'e}rez, Liu, Kahatapitiya, Jia, et~al.]{liu2024mardini}
Haozhe Liu, Shikun Liu, Zijian Zhou, Mengmeng Xu, Yanping Xie, Xiao Han, Juan~C P{\'e}rez, Ding Liu, Kumara Kahatapitiya, Menglin Jia, et~al.
\newblock Mardini: Masked autoregressive diffusion for video generation at scale.
\newblock \emph{arXiv preprint arXiv:2410.20280}, 2024.

\bibitem[Lu et~al.(2023)Lu, Yang, Fei, Huo, Lu, Luo, and Ding]{lu2023vdt}
Haoyu Lu, Guoxing Yang, Nanyi Fei, Yuqi Huo, Zhiwu Lu, Ping Luo, and Mingyu Ding.
\newblock Vdt: General-purpose video diffusion transformers via mask modeling.
\newblock \emph{arXiv preprint arXiv:2305.13311}, 2023.

\bibitem[Ma et~al.(2024)Ma, Wang, Jia, Chen, Liu, Li, Chen, and Qiao]{ma2024latte}
Xin Ma, Yaohui Wang, Gengyun Jia, Xinyuan Chen, Ziwei Liu, Yuan-Fang Li, Cunjian Chen, and Yu Qiao.
\newblock Latte: Latent diffusion transformer for video generation.
\newblock \emph{arXiv preprint arXiv:2401.03048}, 2024.

\bibitem[Mao et~al.(2025)Mao, Zhang, Pan, Jiang, Han, Liu, and Zhou]{mao2025ace}
Chaojie Mao, Jingfeng Zhang, Yulin Pan, Zeyinzi Jiang, Zhen Han, Yu Liu, and Jingren Zhou.
\newblock Ace++: Instruction-based image creation and editing via context-aware content filling.
\newblock \emph{arXiv preprint arXiv:2501.02487}, 2025.

\bibitem[Miao et~al.(2021)Miao, Wei, Wu, Liang, Li, and Yang]{miao2021vspw}
Jiaxu Miao, Yunchao Wei, Yu Wu, Chen Liang, Guangrui Li, and Yi Yang.
\newblock Vspw: A large-scale dataset for video scene parsing in the wild.
\newblock In \emph{Proceedings of the IEEE/CVF conference on computer vision and pattern recognition}, pages 4133--4143, 2021.

\bibitem[Poma et~al.(2020)Poma, Riba, and Sappa]{poma2020dense}
Xavier~Soria Poma, Edgar Riba, and Angel Sappa.
\newblock Dense extreme inception network: Towards a robust cnn model for edge detection.
\newblock In \emph{Proceedings of the IEEE/CVF winter conference on applications of computer vision}, pages 1923--1932, 2020.

\bibitem[Qiu et~al.(2025)Qiu, Fei, Wang, Bai, Yu, Fan, Chen, and Wen]{qiu2025skyreels}
Di Qiu, Zhengcong Fei, Rui Wang, Jialin Bai, Changqian Yu, Mingyuan Fan, Guibin Chen, and Xiang Wen.
\newblock Skyreels-a1: Expressive portrait animation in video diffusion transformers.
\newblock \emph{arXiv preprint arXiv:2502.10841}, 2025.

\bibitem[Qu et~al.(2024)Qu, Zhang, Liu, Wang, Jiang, Gao, Ye, Du, Yuan, and Wu]{qu2024tokenflow}
Liao Qu, Huichao Zhang, Yiheng Liu, Xu Wang, Yi Jiang, Yiming Gao, Hu Ye, Daniel~K Du, Zehuan Yuan, and Xinglong Wu.
\newblock Tokenflow: Unified image tokenizer for multimodal understanding and generation.
\newblock \emph{arXiv preprint arXiv:2412.03069}, 2024.

\bibitem[Ronneberger et~al.(2015)Ronneberger, Fischer, and Brox]{ronneberger2015u}
Olaf Ronneberger, Philipp Fischer, and Thomas Brox.
\newblock U-net: Convolutional networks for biomedical image segmentation.
\newblock In \emph{Medical image computing and computer-assisted intervention--MICCAI 2015: 18th international conference, Munich, Germany, October 5-9, 2015, proceedings, part III 18}, pages 234--241. Springer, 2015.

\bibitem[Silberman et~al.(2012)Silberman, Hoiem, Kohli, and Fergus]{silberman2012indoor}
Nathan Silberman, Derek Hoiem, Pushmeet Kohli, and Rob Fergus.
\newblock Indoor segmentation and support inference from rgbd images.
\newblock In \emph{European conference on computer vision}, pages 746--760. Springer, 2012.

\bibitem[Sun et~al.(2024)Sun, Chu, Zhang, Wu, Dong, Zang, Xiong, Lin, and Wang]{sun2024x}
Zeyi Sun, Ziyang Chu, Pan Zhang, Tong Wu, Xiaoyi Dong, Yuhang Zang, Yuanjun Xiong, Dahua Lin, and Jiaqi Wang.
\newblock X-prompt: Towards universal in-context image generation in auto-regressive vision language foundation models.
\newblock \emph{arXiv preprint arXiv:2412.01824}, 2024.

\bibitem[Tan et~al.(2024)Tan, Liu, Yang, Xue, and Wang]{tan2024ominicontrol}
Zhenxiong Tan, Songhua Liu, Xingyi Yang, Qiaochu Xue, and Xinchao Wang.
\newblock Ominicontrol: Minimal and universal control for diffusion transformer.
\newblock \emph{arXiv preprint arXiv:2411.15098}, 2024.

\bibitem[Team(2020)]{brown2020language}
OpenAI Team.
\newblock Language models are few-shot learners.
\newblock \emph{Advances in neural information processing systems}, 33:\penalty0 1877--1901, 2020.

\bibitem[Villegas et~al.(2022)Villegas, Babaeizadeh, Kindermans, Moraldo, Zhang, Saffar, Castro, Kunze, and Erhan]{villegas2022phenaki}
Ruben Villegas, Mohammad Babaeizadeh, Pieter-Jan Kindermans, Hernan Moraldo, Han Zhang, Mohammad~Taghi Saffar, Santiago Castro, Julius Kunze, and Dumitru Erhan.
\newblock Phenaki: Variable length video generation from open domain textual description.
\newblock \emph{arXiv preprint arXiv:2210.02399}, 2022.

\bibitem[Wang et~al.(2025)Wang, Ai, Wen, Mao, Xie, Chen, Yu, Zhao, Yang, Zeng, et~al.]{wang2025wan}
Ang Wang, Baole Ai, Bin Wen, Chaojie Mao, Chen-Wei Xie, Di Chen, Feiwu Yu, Haiming Zhao, Jianxiao Yang, Jianyuan Zeng, et~al.
\newblock Wan: Open and advanced large-scale video generative models.
\newblock \emph{arXiv preprint arXiv:2503.20314}, 2025.

\bibitem[Wang et~al.(2023{\natexlab{a}})Wang, Yuan, Chen, Zhang, Wang, and Zhang]{wang2023modelscope}
Jiuniu Wang, Hangjie Yuan, Dayou Chen, Yingya Zhang, Xiang Wang, and Shiwei Zhang.
\newblock Modelscope text-to-video technical report.
\newblock \emph{arXiv preprint arXiv:2308.06571}, 2023{\natexlab{a}}.

\bibitem[Wang et~al.(2023{\natexlab{b}})Wang, Wang, Cao, Shen, and Huang]{wang2023images}
Xinlong Wang, Wen Wang, Yue Cao, Chunhua Shen, and Tiejun Huang.
\newblock Images speak in images: A generalist painter for in-context visual learning.
\newblock In \emph{Proceedings of the IEEE/CVF Conference on Computer Vision and Pattern Recognition}, pages 6830--6839, 2023{\natexlab{b}}.

\bibitem[Wang et~al.(2024)Wang, Xia, Chen, Yu, Wang, Gong, and Liu]{wang2024lavin}
Zhaoqing Wang, Xiaobo Xia, Runnan Chen, Dongdong Yu, Changhu Wang, Mingming Gong, and Tongliang Liu.
\newblock Lavin-dit: Large vision diffusion transformer.
\newblock \emph{arXiv preprint arXiv:2411.11505}, 2024.

\bibitem[Xie et~al.(2024)Xie, Mao, Bai, Zhang, Wang, Lin, Gu, Chen, Yang, and Shou]{xie2024show}
Jinheng Xie, Weijia Mao, Zechen Bai, David~Junhao Zhang, Weihao Wang, Kevin~Qinghong Lin, Yuchao Gu, Zhijie Chen, Zhenheng Yang, and Mike~Zheng Shou.
\newblock Show-o: One single transformer to unify multimodal understanding and generation.
\newblock \emph{arXiv preprint arXiv:2408.12528}, 2024.

\bibitem[Yan et~al.(2021)Yan, Zhang, Abbeel, and Srinivas]{yan2021videogpt}
Wilson Yan, Yunzhi Zhang, Pieter Abbeel, and Aravind Srinivas.
\newblock Videogpt: Video generation using vq-vae and transformers.
\newblock \emph{arXiv preprint arXiv:2104.10157}, 2021.

\bibitem[Yang et~al.(2024)Yang, Teng, Zheng, Ding, Huang, Xu, Yang, Hong, Zhang, Feng, et~al.]{yang2024cogvideox}
Zhuoyi Yang, Jiayan Teng, Wendi Zheng, Ming Ding, Shiyu Huang, Jiazheng Xu, Yuanming Yang, Wenyi Hong, Xiaohan Zhang, Guanyu Feng, et~al.
\newblock Cogvideox: Text-to-video diffusion models with an expert transformer.
\newblock \emph{arXiv preprint arXiv:2408.06072}, 2024.

\bibitem[Zhang et~al.(2025)Zhang, Yuan, Song, Wang, and Liu]{zhang2025easycontrol}
Yuxuan Zhang, Yirui Yuan, Yiren Song, Haofan Wang, and Jiaming Liu.
\newblock Easycontrol: Adding efficient and flexible control for diffusion transformer.
\newblock \emph{arXiv preprint arXiv:2503.07027}, 2025.

\bibitem[Zhou et~al.(2019)Zhou, Zhao, Puig, Xiao, Fidler, Barriuso, and Torralba]{zhou2019semantic}
Bolei Zhou, Hang Zhao, Xavier Puig, Tete Xiao, Sanja Fidler, Adela Barriuso, and Antonio Torralba.
\newblock Semantic understanding of scenes through the ade20k dataset.
\newblock \emph{International Journal of Computer Vision}, 127\penalty0 (3):\penalty0 302--321, 2019.

\bibitem[Zi et~al.(2025)Zi, Ruan, Chen, Qi, Hao, Zhao, Huang, Liang, Xiao, and Wong]{zi2025senorita}
Bojia Zi, Penghui Ruan, Marco Chen, Xianbiao Qi, Shaozhe Hao, Shihao Zhao, Youze Huang, Bin Liang, Rong Xiao, and Kam-Fai Wong.
\newblock Se\~norita-2m: A high-quality instruction-based dataset for general video editing by video specialists.
\newblock \emph{arXiv preprint arXiv:2502.06734}, 2025.

\end{thebibliography}
}
\clearpage
\appendix
\section{Implementation Details}
\label{sec:details}
\Paragraph{Data Collection.} We collect paired data across six representative tasks, including generation tasks (1) scribble map transfer, (2) Van Gogh style transfer and (3) camera movement transfer and perception tasks (4) depth map prediction, (5) semantic segmentation prediction, and (6) salient object tracking.
For task (1)(4)(6), we collect the source clips ($A, B$) from the Se\~norita-2M dataset~\cite{zi2025senorita}, and the annotated clips ($A', B'$) are obtained using preprocessing tools from VACE~\cite{vace} code repository\footnote{\url{https://github.com/ali-vilab/VACE}}.
For task (2)(3), we use the source videos from Se\~norita-2M dataset~\cite{zi2025senorita} and edit them using TokenFlow~\cite{qu2024tokenflow} and computer software CapCut\footnote{\url{https://www.capcut.com/}}, respectively.
For task (5), we source data from VSPW dataset~\cite{miao2021vspw}.

\Paragraph{Fine-tuning Details.}
We employ Wan2.1-T2V-1.3B\footnote{\url{https://github.com/Wan-Video/Wan2.1}} as the backbone video generation model for our experiments.
The model is trained for $20$\textasciitilde$40$ epochs for each vision tasks, with each epoch consisting of $200$ iterations.
For co-training across all vision tasks and contexts, we train the model for $20$ epochs, with 1,200 iterations per epoch.
Training on two A800 GPUs, each epoch takes about $12$ minutes for $200$ iterations and roughly one hour for $1200$ iterations.

\Paragraph{Experimental Details.}
In fine-tuning with mixed vision contexts, we randomly choose the vision contexts of each training sample.
For tasks (1)(2)(4)(5)(6), we sample vision contexts \uppercase\expandafter{\romannumeral1} and \uppercase\expandafter{\romannumeral2} each with probability $p=0.3$, context \uppercase\expandafter{\romannumeral3} with $p=0.4$.
For task (3), since the transformation pertains to the temporal dimension, sampling is limited to contexts \uppercase\expandafter{\romannumeral1} and \uppercase\expandafter{\romannumeral4}, each with probability $p=0.5$.
In the ablation study investigating the impact of text, we utilize prompts at multiple levels of granularity, including detailed, rough, and null texts.
The prompt template is illustrated in~\figref{prompt_template}.
For all other experiments, we consistently use the detailed text prompt.

%
%
%
%
\section{Comparison between LVM and video generation model}

As summarized in ~\tabref{cost}, the training data required by LVM~\cite{bai2024sequential} is complex to construct, while the video generation model Wan is pretrained only on raw images and videos.
Although Wan uses more total training tokens than LVM, it achieves higher visual quality by employing an 8× downsampling encoder, compared to LVM's 16× downsampling.
Additionally, Wan has fewer parameters than the released version of LVM, resulting in lower computational costs.

\begin{figure}[!t] 
\vspace{-1 em}
\centering
\begin{AIbox}{Text Prompts}
{\color{deepblue}\bf Detailed Text:} ``[clip1] is the original source video, and [clip2] is its corresponding segmentation map. [clip3] is another, different source video. In [clip4], the segmentation map transformation applied from [clip1] to [clip2] is similarly applied to [clip3].''\\

{\color{deepblue}\bf Rough Text:} ``[clip1] and [clip2] form an editing pair. Apply the same transformation observed from [clip1] to [clip2] to [clip3], and generate [clip4].'' \\

{\color{deepblue}\bf Null Text:} ``''
\end{AIbox} 
\vspace{-5 mm}
\caption{\textbf{Text Prompts at multiple levels of granularity.}}
\label{fig:prompt_template}
\vspace{-2 mm}
\end{figure}
\begin{table}[t]
\centering
\begin{tabular}{>{\centering\arraybackslash}p{1.7cm}>{\centering\arraybackslash}p{3.7cm}>{\centering\arraybackslash}p{1.6cm}}
\toprule
& \textbf{LVM} & \textbf{Wan} \\
\midrule
\parbox{1.7cm}{\centering Dataset \\ Composition} & 
\begin{tabular}[t]{@{}l@{}}
1. Single images \\ 
2. Image sequences \\
3. Images with annotations \\
4. Image sequences with \\
annotations
\end{tabular}
& Raw images and videos \\
\midrule
\parbox{1.7cm}{\centering Dataset \\ Scale} & $420$B tokens & $\mathcal{O}(1)$T tokens~\cite{wang2025wan} \\
\midrule
\parbox{1.7cm}{\centering Downsample \\ Ratio} & $16X$ & $8X$ \\
\midrule
Parameters & $7$B (released version) & $1.3$B \\
\bottomrule
\end{tabular}
\vspace{-2 mm}
\caption{\textbf{Comparison between LVM~\cite{bai2024sequential} and the video generation model Wan~\cite{wang2025wan}.}}
\label{tab:cost}
\vspace{-6 mm}
\end{table}
\section{Additional Experimental Results}
We provide additional results related to four experiments presented in the main paper.
Specifically, we show the results of each vision task across all contexts in~\subsecref{context}, the performance of each task under mixed-context fine-tuning in~\subsecref{mixed},
the impact of text prompts in~\subsecref{impact}, and the results of  co-training with all tasks and contexts in~\subsecref{co-train}.
%
\begin{figure*}[t!]
    \centering
\includegraphics[width=1\linewidth]{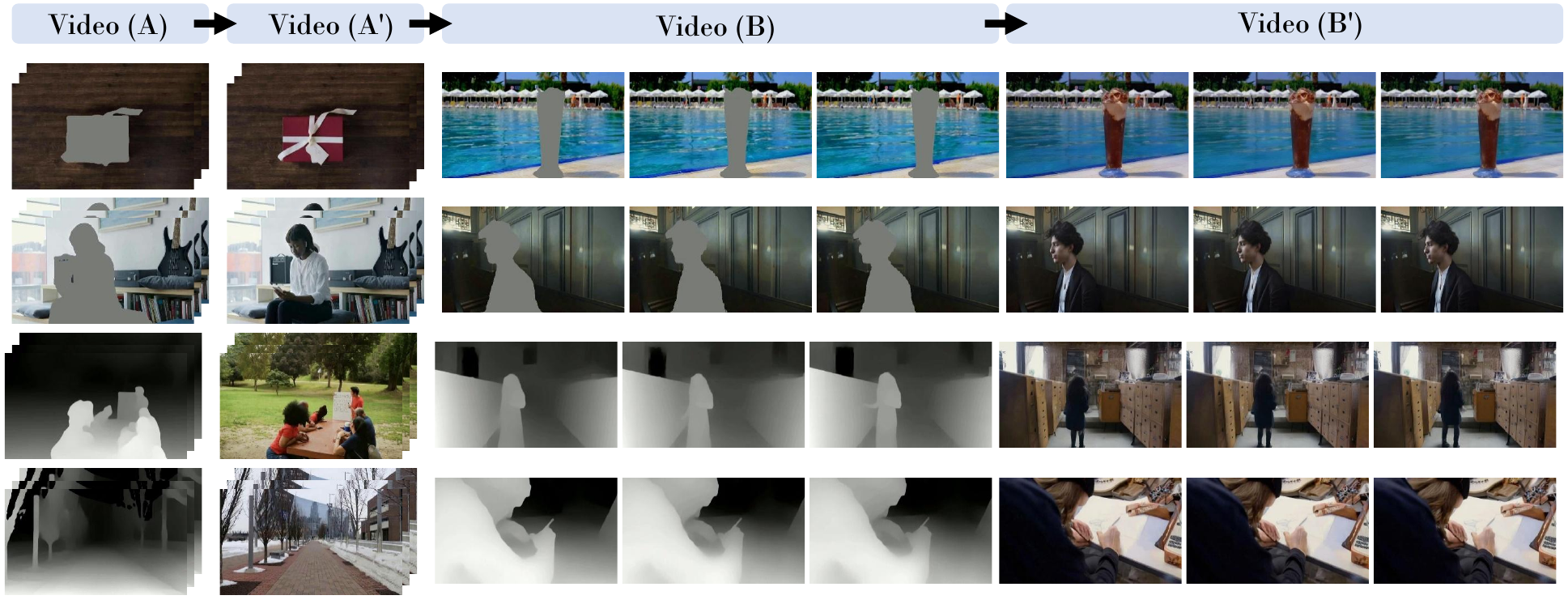}  
    \vspace{-7 mm}
  \caption{\textbf{Results of conditional generation tasks.}
    %
    }
    \vspace{-1 mm}
    \label{fig:cond_supp}
\end{figure*}
\begin{figure*}[t!]
    \centering
\includegraphics[width=1\linewidth]{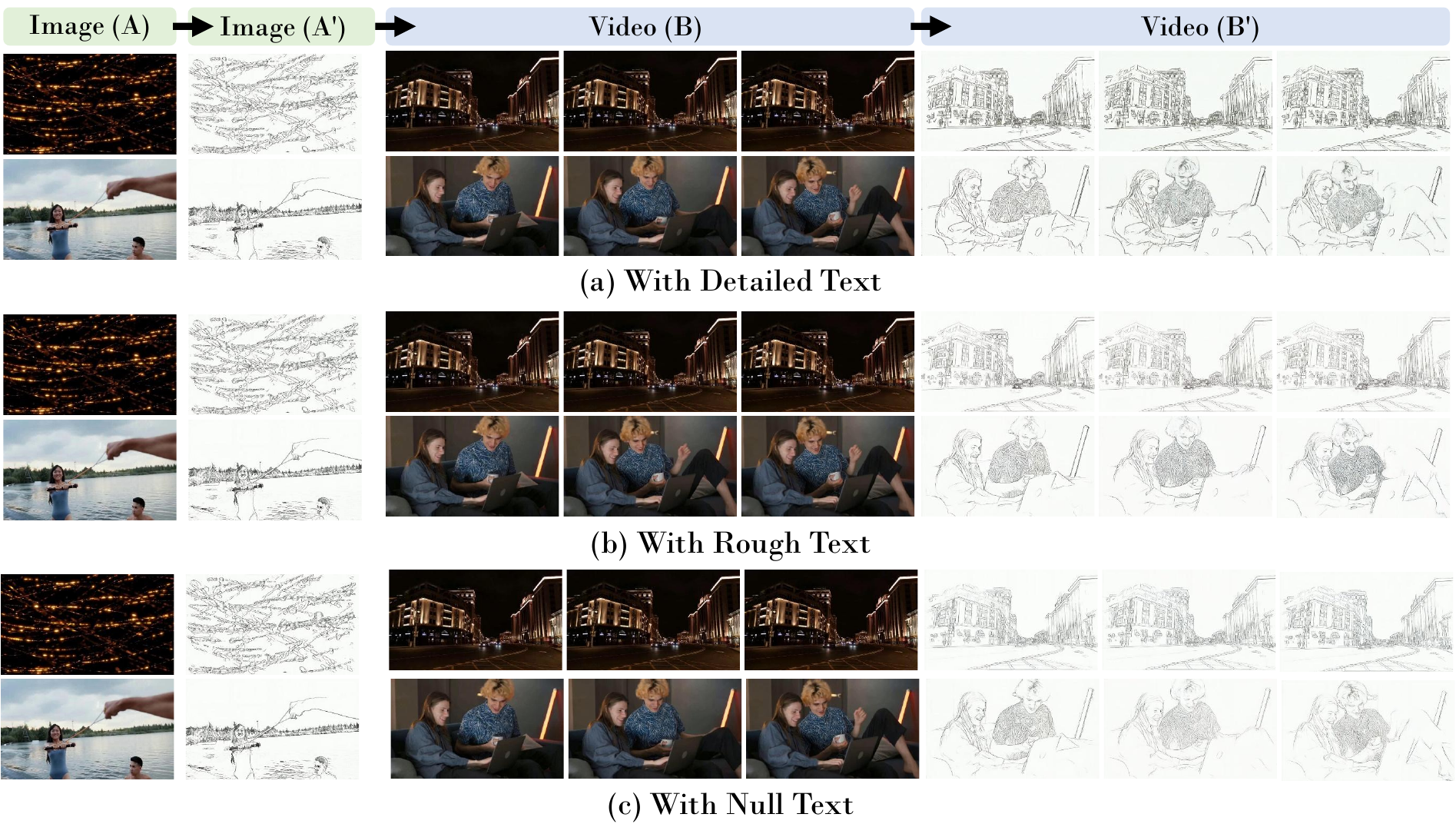}  
  \caption{\textbf{Impact of texts under contexts \uppercase\expandafter{\romannumeral3}.}
    %
    }
    \vspace{-3 mm}
    \label{fig:i_v}
\end{figure*}

%
\subsection{Conditional Generation Tasks}
\label{subsec:cond}
As shown in~\figref{cond_supp}, the video generation model is capable of performing conditional generation tasks based on depth maps or masked videos.

\subsection{Performance Across Different Contexts}
\label{subsec:context}
\begin{figure*}[t!]
    \centering
\includegraphics[width=1\linewidth]{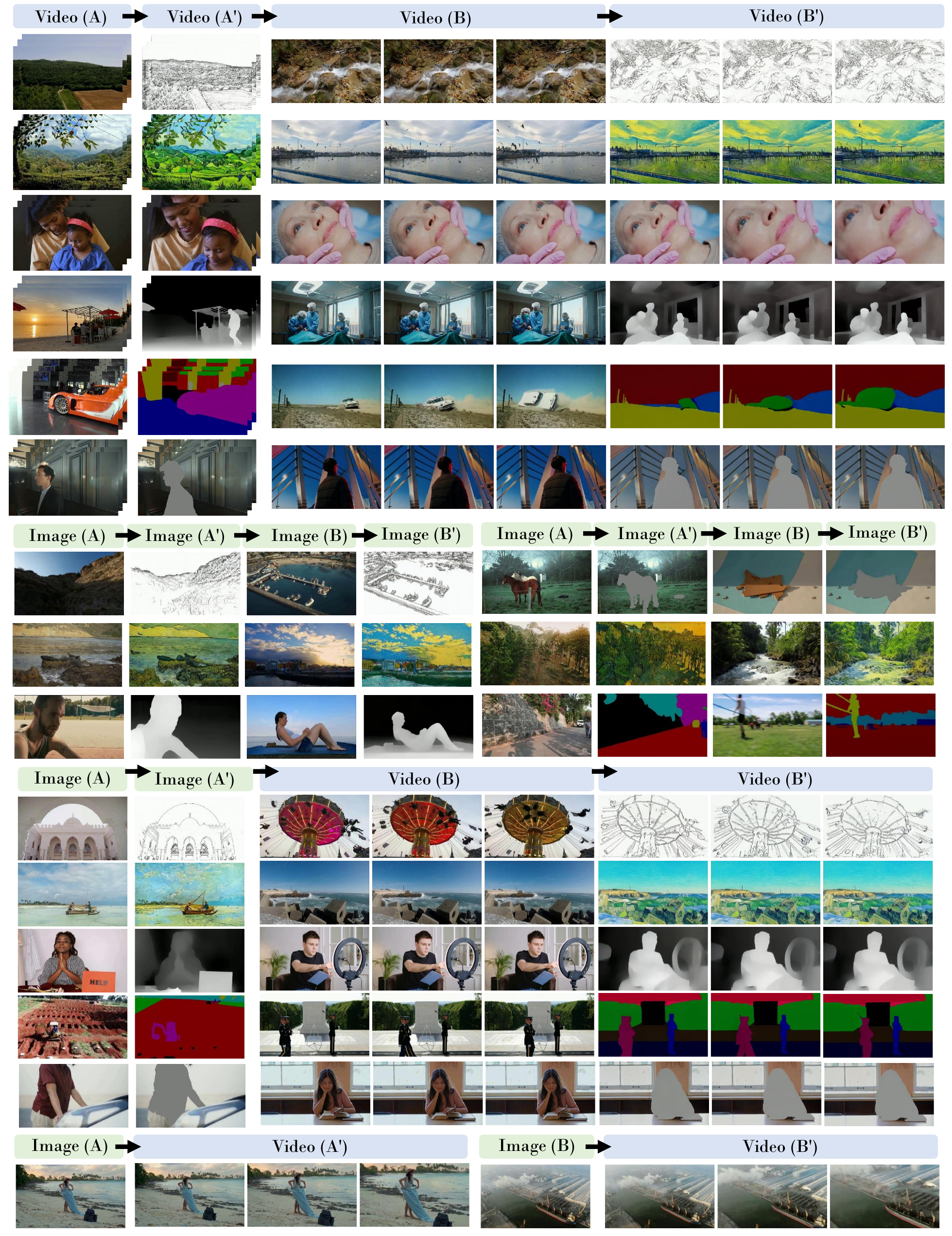}  
    \vspace{-6 mm}
    \caption{\textbf{Additional results of various vision tasks across context types.}
    %
    }
    \vspace{-5 mm}
    \label{fig:more_results}
\end{figure*}

We present the performance of each vision task across different contexts in~\figref{more_results}.
The results demonstrate that the fine-tuned video generation model effectively handles not only image and video tasks, but also cross-modal and cross-data-source tasks.
%

\subsection{Performance under Mixed-Context Fine-Tuning}
\label{subsec:mixed}
\begin{figure*}[t!]
    \centering
\includegraphics[width=1\linewidth]{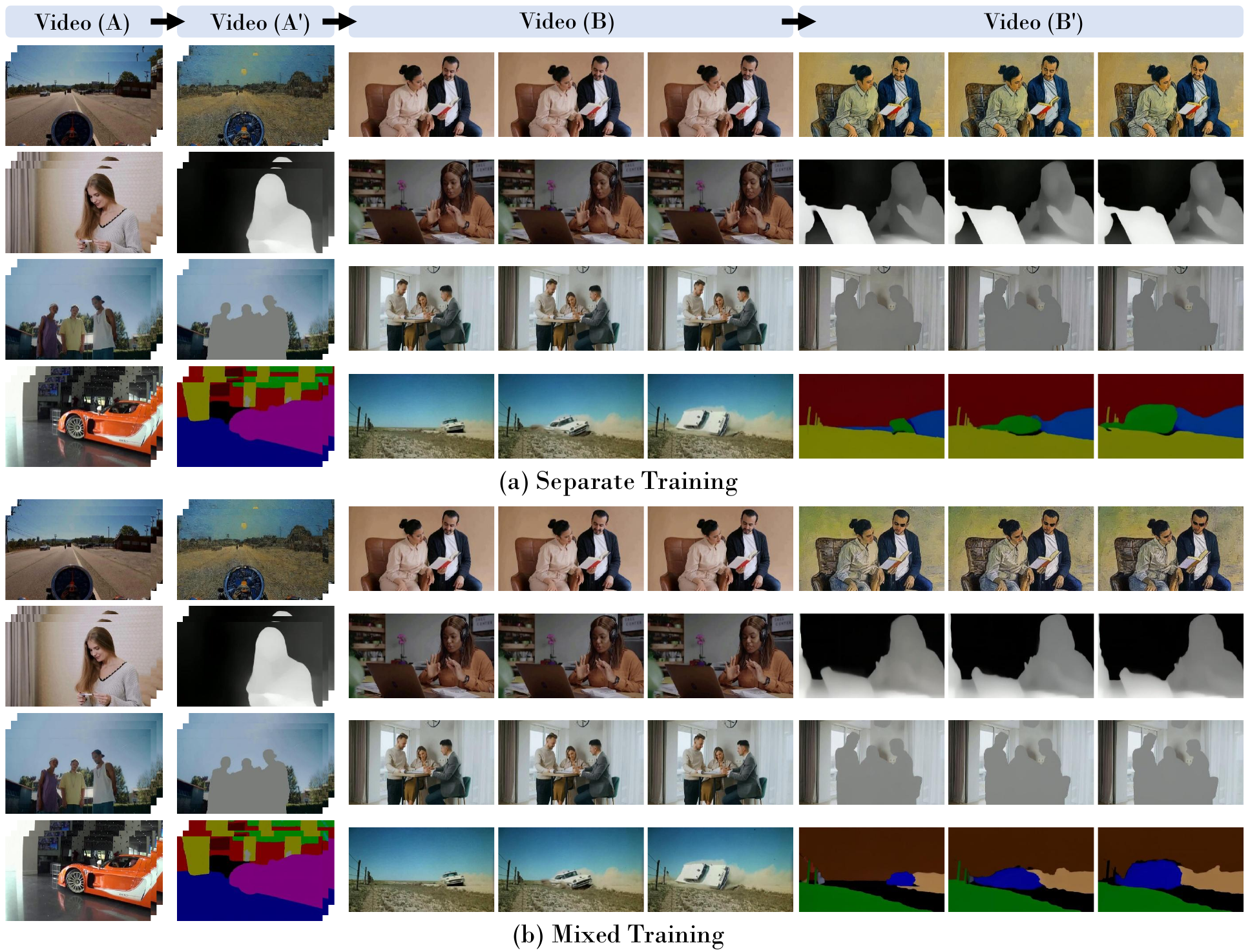}  
    \vspace{-7 mm}
    \caption{\textbf{Performance under separate and mixed training for context \uppercase\expandafter{\romannumeral1}.}
    %
    }
    \vspace{-3 mm}
    \label{fig:mix_train1}
\end{figure*}

\begin{figure*}[t!]
    \centering
\includegraphics[width=1\linewidth]{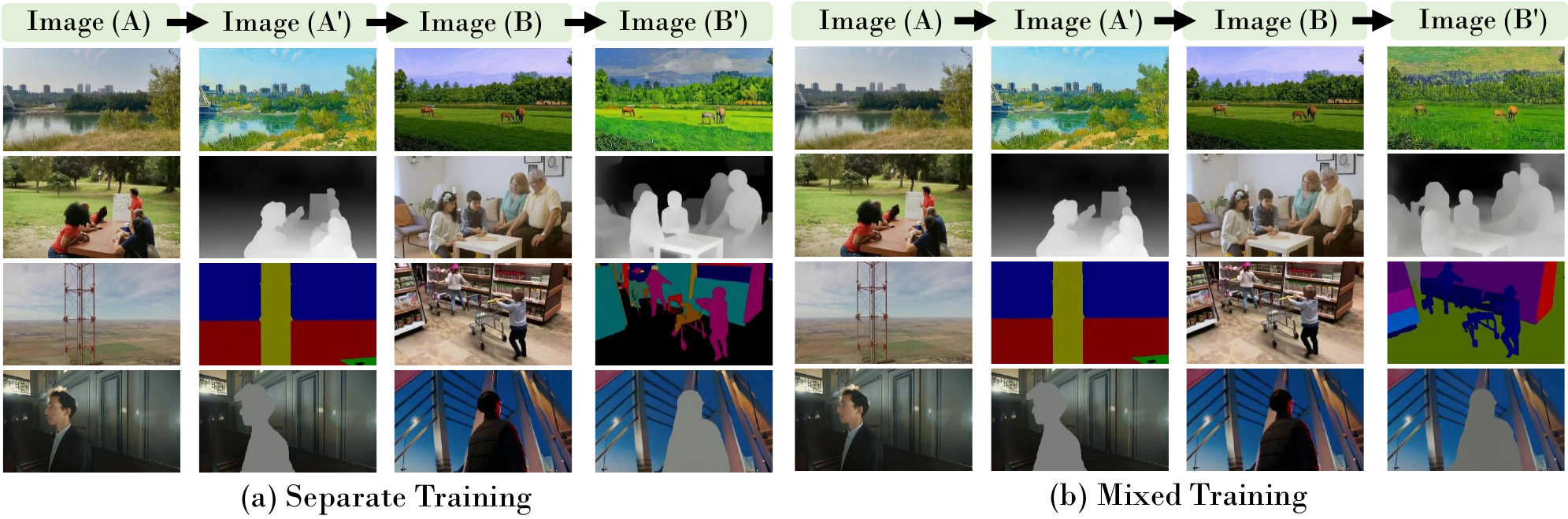}  
    \caption{\textbf{Performance under separate and mixed training for context \uppercase\expandafter{\romannumeral2}.}
    %
    }
    \vspace{-5 mm}
    \label{fig:mix_train2}
\end{figure*}

\begin{figure*}[t!]
    \centering
\includegraphics[width=1\linewidth]{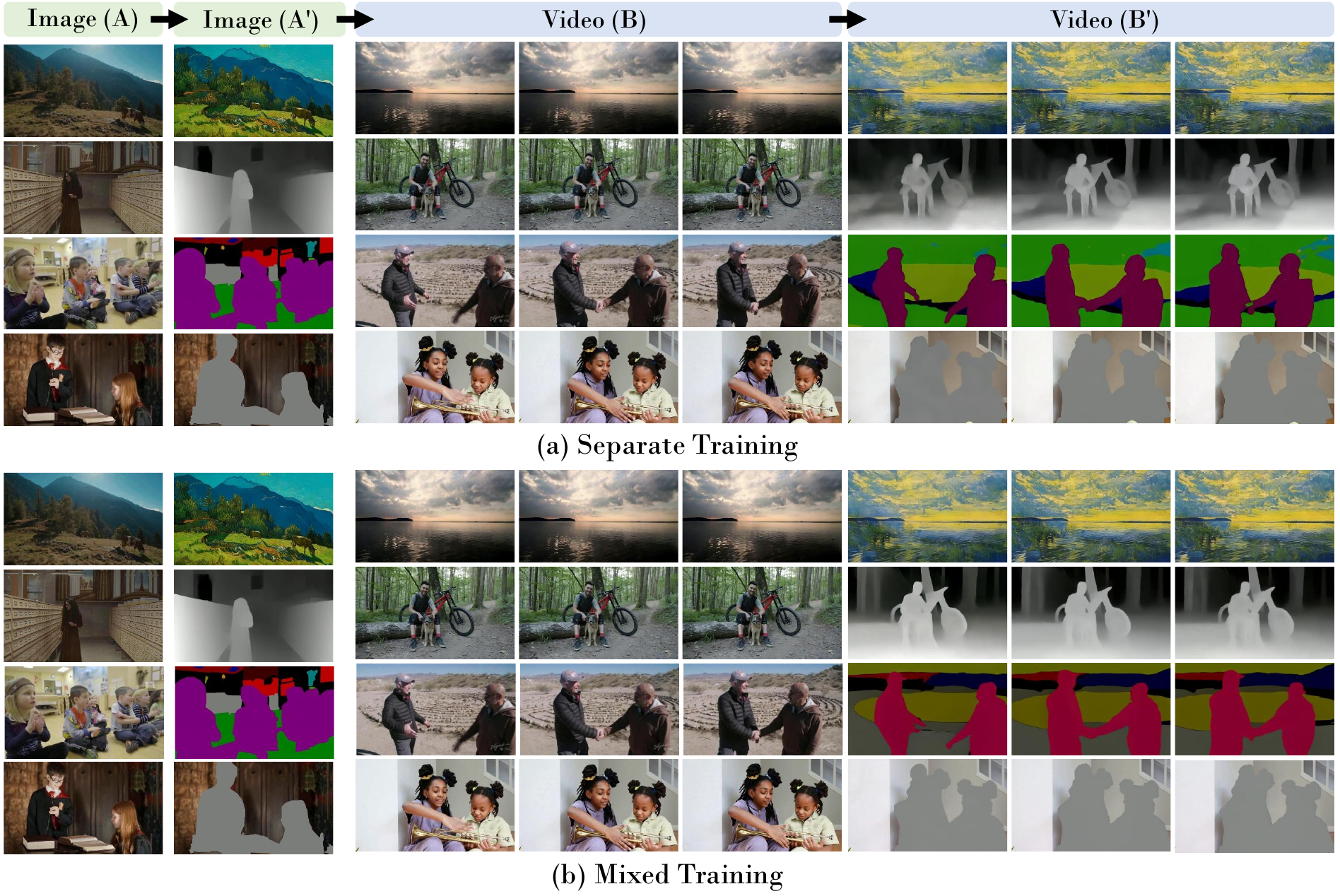}  
    \vspace{-7 mm}
    \caption{\textbf{Performance under separate and mixed training for context \uppercase\expandafter{\romannumeral3}.}
    %
    }
    \vspace{-3 mm}
    \label{fig:mix_train3}
\end{figure*}

As shown in~\figref{mix_train1},~\figref{mix_train2} and~\figref{mix_train3} ,  when fine-tuned on mixed vision contexts, the model can automatically adjust to each vision context, demonstrating strong generalization ability.

\subsection{Co-training with All Vision Tasks and Contexts}
\label{subsec:co-train}
\begin{figure*}[t!]
    \centering
\includegraphics[width=1\linewidth]{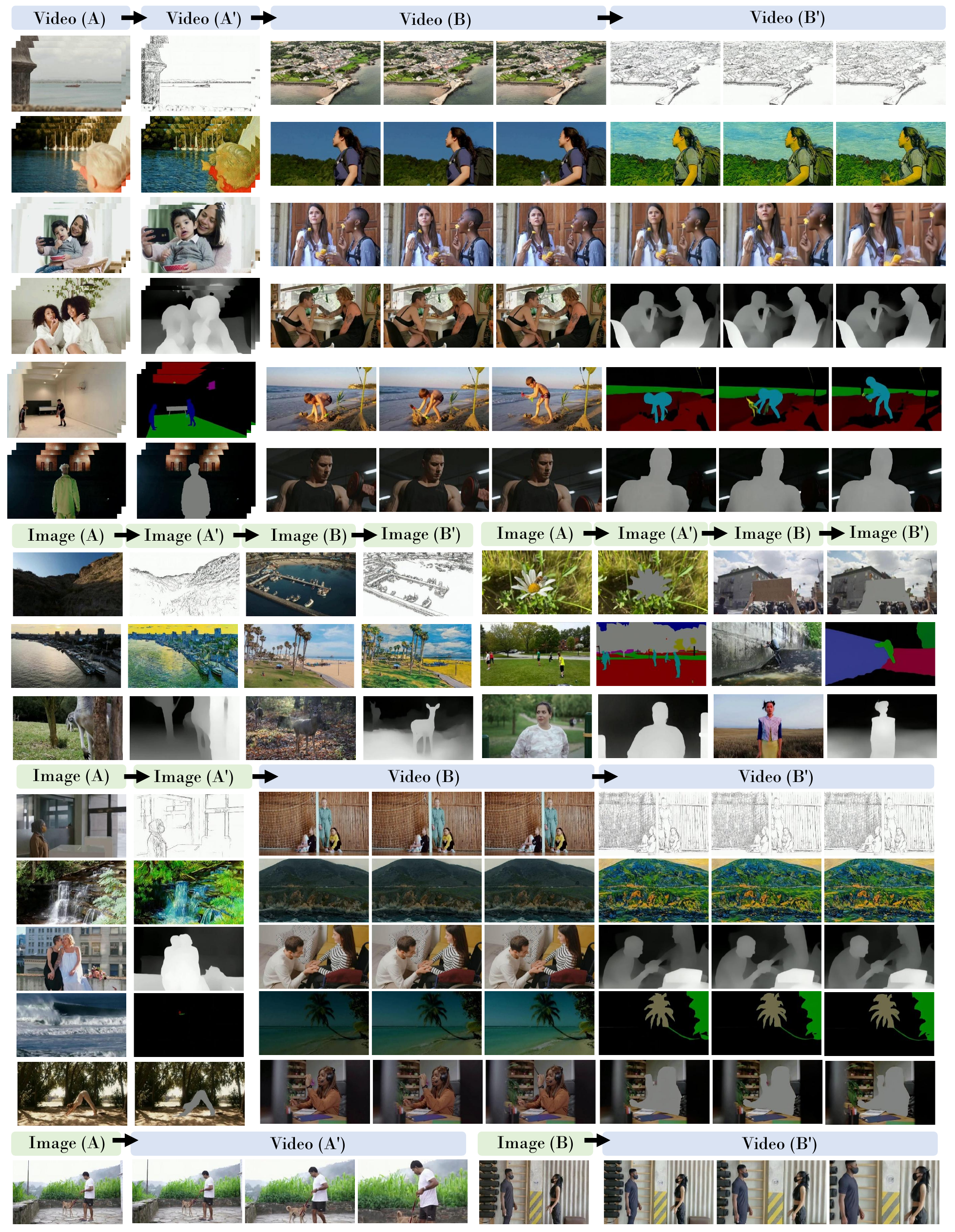}  
    \vspace{-6 mm}
  \caption{\textbf{Co-training with all vision tasks and contexts.}
    %
    }
    \label{fig:co-train-supp}
\end{figure*}

As shown in~\figref{co-train-supp}, when co-trained on all vision tasks under mixed contexts, the model achieves consistent performance across tasks and contexts, demonstrating robust generalization ability with limited supervision.
%

\subsection{Impact of Texts Across Different Vision Contexts}
\label{subsec:impact}
\begin{figure*}[t!]
    \centering
\includegraphics[width=1\linewidth]{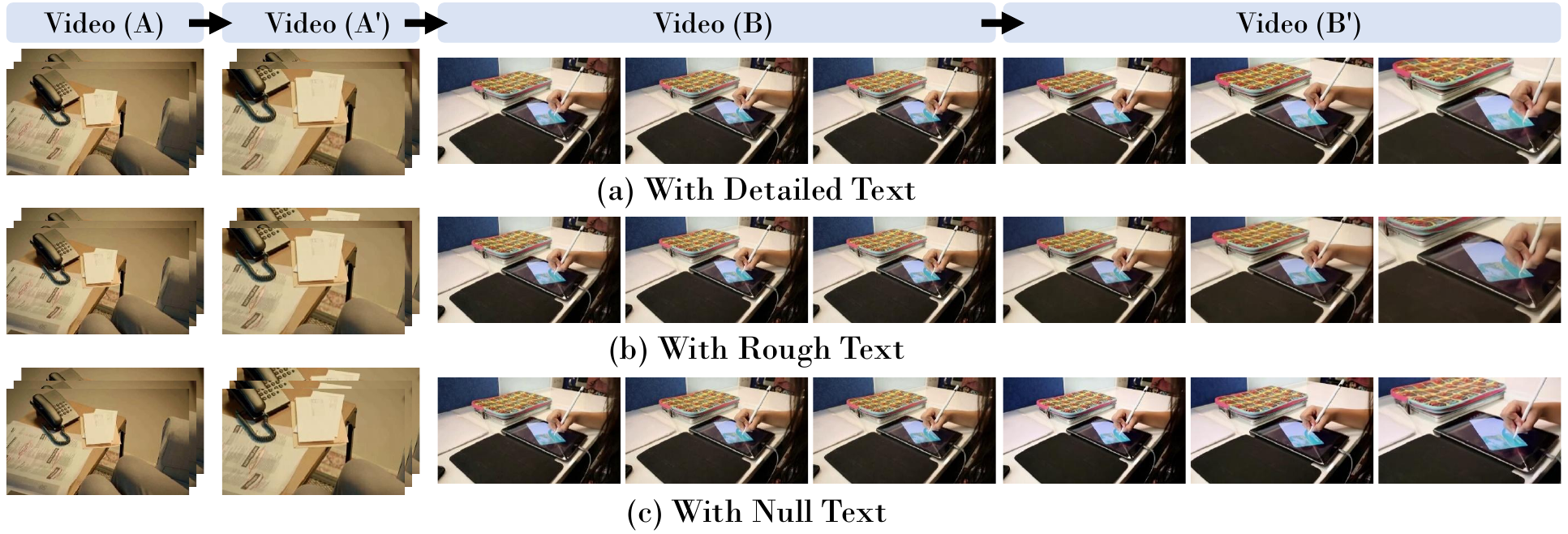}  
  \caption{\textbf{Impact of texts under contexts \uppercase\expandafter{\romannumeral1}.}
    %
    }
    \vspace{-3 mm}
    \label{fig:v_v}
\end{figure*}

\begin{figure*}[t!]
    \centering
\includegraphics[width=1\linewidth]{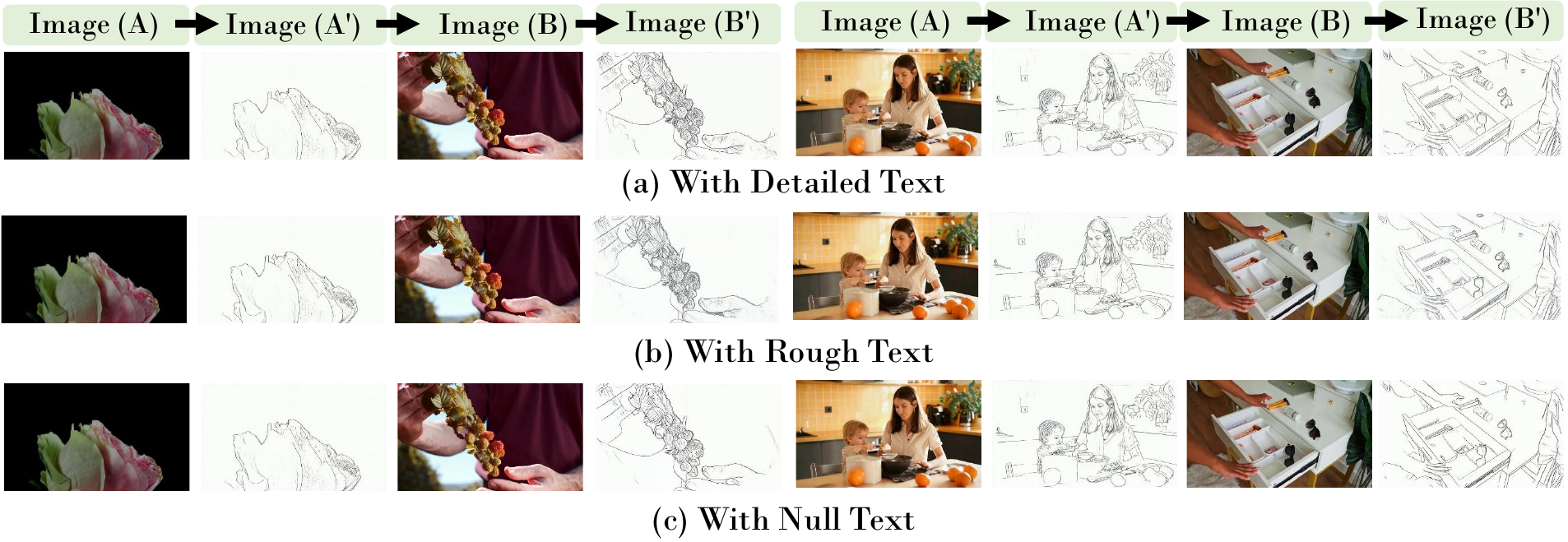}  
  \caption{\textbf{Impact of texts under contexts \uppercase\expandafter{\romannumeral2}.}
    %
    }
    \vspace{-3 mm}
    \label{fig:i_i}
\end{figure*}

\begin{figure*}[t!]
    \centering
\includegraphics[width=1\linewidth]{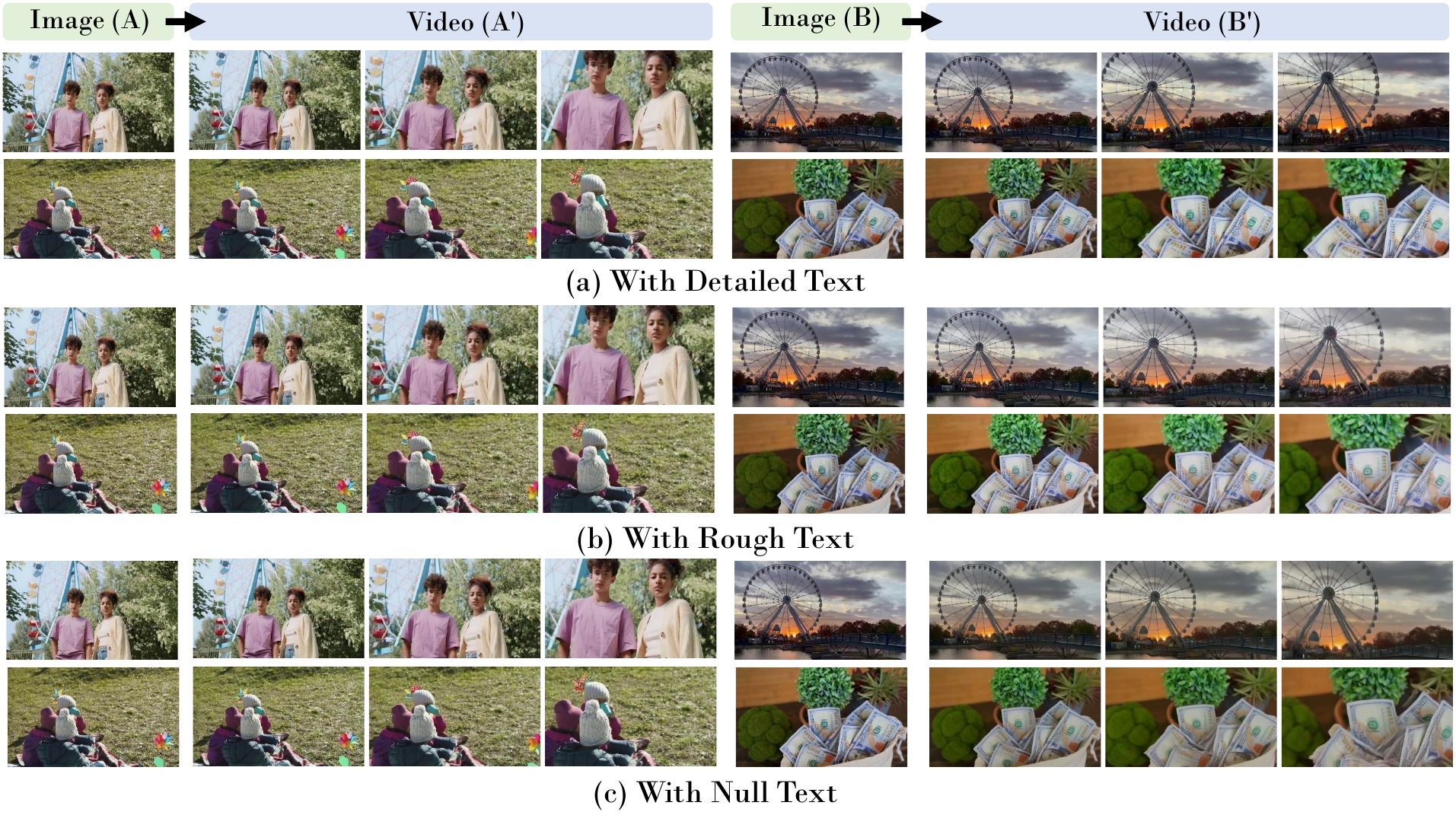}  
  \caption{\textbf{Impact of texts under contexts \uppercase\expandafter{\romannumeral4}.}
    %
    }
    \vspace{-3 mm}
    \label{fig:i_v_i_v}
\end{figure*}

As shown in~\figref{v_v}, ~\figref{i_i}, \figref{i_v}, and~\figref{i_v_i_v}, the model effectively learns the relationships among the four clips across different vision contexts without explicit textual guidance, demonstrating strong in-context learning capabilities in the temporal dimension.



\end{document}